\documentclass[pmlr]{jmlr}

\usepackage{longtable}

\usepackage{booktabs}
\usepackage{multirow}
\usepackage[load-configurations=version-1]{siunitx} 

\makeatletter
\def\set@curr@file#1{\def\@curr@file{#1}} 
\makeatother

\usepackage{dsfont}
\usepackage{todonotes}
\usepackage{wrapfig}
\usepackage{cleveref}

\usepackage{xcolor}
\usepackage{hyperref}
\hypersetup{
    colorlinks,
    linkcolor={red!50!black},
    citecolor={blue!50!black},
    urlcolor={blue!80!black}
}

\usepackage{listings}
\definecolor{codegreen}{rgb}{0,0.6,0}
\definecolor{codegray}{rgb}{0.5,0.5,0.5}
\definecolor{codepurple}{rgb}{0.58,0,0.82}
\definecolor{backcolour}{rgb}{0.95,0.95,0.92}
\lstdefinestyle{mystyle}{
    backgroundcolor=\color{backcolour},   
    commentstyle=\color{codegreen},
    keywordstyle=\color{magenta},
    numberstyle=\tiny\color{codegray},
    stringstyle=\color{codepurple},
    basicstyle=\ttfamily\footnotesize,
    breakatwhitespace=false,         
    breaklines=true,                 
    captionpos=b,                    
    keepspaces=true,                 
    numbersep=5pt,                  
    showspaces=false,                
    showstringspaces=false,
    showtabs=false,                  
    tabsize=2
}
\jmlrvolume{}
\jmlryear{}
\jmlrworkshop{}

\usepackage{xparse}
\usepackage{bm,upgreek}

\ExplSyntaxOn
\NewDocumentCommand\vc{m}
 {
  \commexo_vector:n { #1 }
 }

\cs_new_protected:Npn \commexo_vector:n #1
 {
  \tl_map_inline:nn { #1 }
   {
    \commexo_vector_inner:n { ##1 }
   }
 }

\cs_new_protected:Npn \commexo_vector_inner:n #1
 {
  \tl_if_in:VnTF \g_commexo_latin_tl { #1 }
   {
    \mathbf { #1 } 
   }
   {
    \tl_if_in:VnTF \g_commexo_ucgreek_tl { #1 }
     {
      \bm { #1 } 
     }
     {
      \tl_if_in:VnTF \g_commexo_lcgreek_tl { #1 }
       {
        \commexo_makeboldupright:n { #1 }
       }
       {
        #1 
       }
     }
   }
 }

\cs_new_protected:Npn \commexo_makeboldupright:n #1
 {
  \bm { \use:c { up \cs_to_str:N #1 } }
 }

\tl_new:N \g_commexo_latin_tl
\tl_new:N \g_commexo_ucgreek_tl
\tl_new:N \g_commexo_lcgreek_tl
\tl_gset:Nn \g_commexo_latin_tl
 {
  ABCDEFGHIJKLMNOPQRSTUVWXYZ
  abcdefghijklmnopqrstuvwxyz
 }
\tl_gset:Nn \g_commexo_ucgreek_tl
 {
  \Gamma\Delta\Theta\Lambda\Pi\Sigma\Upsilon\Phi\Chi\Psi\Omega
 }
\tl_gset:Nn \g_commexo_lcgreek_tl
 {
  \alpha\beta\gamma\delta\epsilon\zeta\eta\theta\iota\kappa
  \lambda\mu\nu\xi\pi\rho\sigma\tau\upsilon\phi\chi\psi\omega
  \varepsilon\vartheta\varpi\varphi\varsigma\varrho
 }

\ExplSyntaxOff

 \newcounter{mycomment}

\title[Model-based metrics]{Model-based metrics: Sample-efficient estimates of predictive model subpopulation performance}

\author{\Name{Andrew C.~Miller} \Email{acmiller@apple.com}
     \addr \\
     Apple
     \AND
     \Name{Leon A.~Gatys} \Email{lgatys@apple.com}
     \addr \\
     Apple
     \AND
     \Name{Joseph Futoma} \Email{jfutoma@apple.com}
     \addr \\
     Apple 
     \AND
     \Name{Emily B.~Fox} \Email{emily\_fox@apple.com}
     \addr \\
     Apple 
}
\date{}

\begin{document}

\maketitle
\begin{abstract}
Machine learning models --- now commonly developed to screen, diagnose, or predict health conditions --- are evaluated with a variety of performance metrics.
An important first step in assessing the practical utility of a model is to evaluate its average performance over an entire population of interest.
In many settings, it is also critical that the model makes good predictions within predefined subpopulations.
For instance, showing that a model is fair or equitable requires evaluating the model's performance in different demographic subgroups. 
However, subpopulation performance metrics are typically computed using only data from that subgroup, resulting in higher variance estimates for smaller groups. 
We devise a procedure to measure subpopulation performance that can be more sample-efficient than the typical subsample estimates. 
We propose using an \emph{evaluation model} --- a model that describes the conditional distribution of the predictive model score --- to form \emph{model-based metric} (MBM) estimates. 
Our procedure incorporates model checking and validation, and we propose a computationally efficient approximation of the traditional nonparametric bootstrap to form confidence intervals.  
We evaluate MBMs on two main tasks: a semi-synthetic setting where ground truth metrics are available and a real-world hospital readmission prediction task.
We find that MBMs consistently produce more accurate and lower variance estimates of model performance for small subpopulations. 
\end{abstract}

\section{Introduction}
Machine learning (ML) is increasingly used to screen, diagnose, and predict health conditions. 
As of March 2020, at least 222 medical devices using ML were approved by the US Food and Drug administration \citep{muehlematter2021approval}, spanning medical specialties, including radiology, cardiology, and ophthalmology.
Ensuring that a model is suitable for deployment in a real-world setting requires quantifying its performance using evaluation metrics such as the area under the receiver operating characteristic curve (AUC), positive predictive value (PPV), and false positive rate (FPR); robustly estimating model performance is of particular importance in medical contexts.
It is also important to assess a model's performance \emph{within} subpopulations of interest, which is a necessary step in diagnosing issues with the fairness or equity of model predictions, and to identify avenues for model improvement.
The typical way to estimate the performance of a model on subgroups is to compute the relevant set of metrics on the subpopulation \emph{subsample}.
However, subsample-based estimates of model performance are more uncertain in smaller subpopulations.
This becomes especially problematic when we examine groups defined by the intersection of multiple demographic categories (e.g., sex \emph{and} age). And in clinical studies it can be challenging to gather more data from such underrepresented populations. 

Consider a hypothetical ML model to predict future cardiac events with the goal of screening individuals into ``low'' and ``elevated'' risk categories defined by the Framingham cardiovascular disease risk score \citep{d2008general}. 
To scrutinize the accuracy and fairness of this model, we measure its performance on demographic subpopulations and form estimates of metrics such as AUC, PPV, and FPR within each group.
Suppose that our validation cohort consists of only a few thousand individuals, a modest sample size that may occur in many different medical applications.
If our data is representative of the entire United States, we would only expect about 1.3\% of individuals to be non-Hispanic African American males between 39 and 52 years old.
In our observed sample, this would likely correspond to a few dozen individuals at most.
At such a small sample size, our estimates of various model metrics for this group will be extremely imprecise, impeding our ability to improve the model's performance within this population.

The crux of the problem is simple: \emph{restricting estimates to subpopulation subsamples does not use all the information available}.
Individuals with similar, but not exactly the same, demographic labels may contain information that can improve subpopulation estimates.
For instance, in our cardiac risk application, it may be that non-Hispanic Caucasian males between 39 and 52, or non-Hispanic African American males between 26 and 39, contain helpful information about the non-Hispanic African American males between 39 to 52 subgroup.
This is exactly the benefit of multi-level modeling: share information between groups in a principled manner to improve the estimate within each group \citep{gelman2013bayesian}.

\begin{figure*}[t]
\floatconts
  {fig:subfigex}
  {
  \vspace{-1.5em}
  \caption{
      \small{Method overview: (a) For subpopulations determined by age and Fitzpatrick skin type (FP) bins, we evaluate model performance \emph{within} each subpopulation.  Some subpopulations are more populous (green) than others (gold).  Less data leads to noisier metric estimates.
      (b) We propose model-based metric estimators, which relies on a joint estimate of the class conditionals informed by all observations.  The model incorporates more information into metric estimates for small subpopulations. For example, the subpopulation outlined in gold will be informed by other observations with the same skin type (FP-2) and the same age bin (65-80). This leads to more sample-efficient and informative estimates of model performance.
      (c) Subpopulation frequencies, defined by race and age categories, in the National Health and Nutrition Examination Survey, designed to be representative of the United States population.  Some subpopulations are over 10$\times$ larger than others.
      }
  }}
  {%
    \subfigure[\small{Subpopulation samples}]{
      \label{fig:example-subpopulation}
      \includegraphics[height=1.7in]{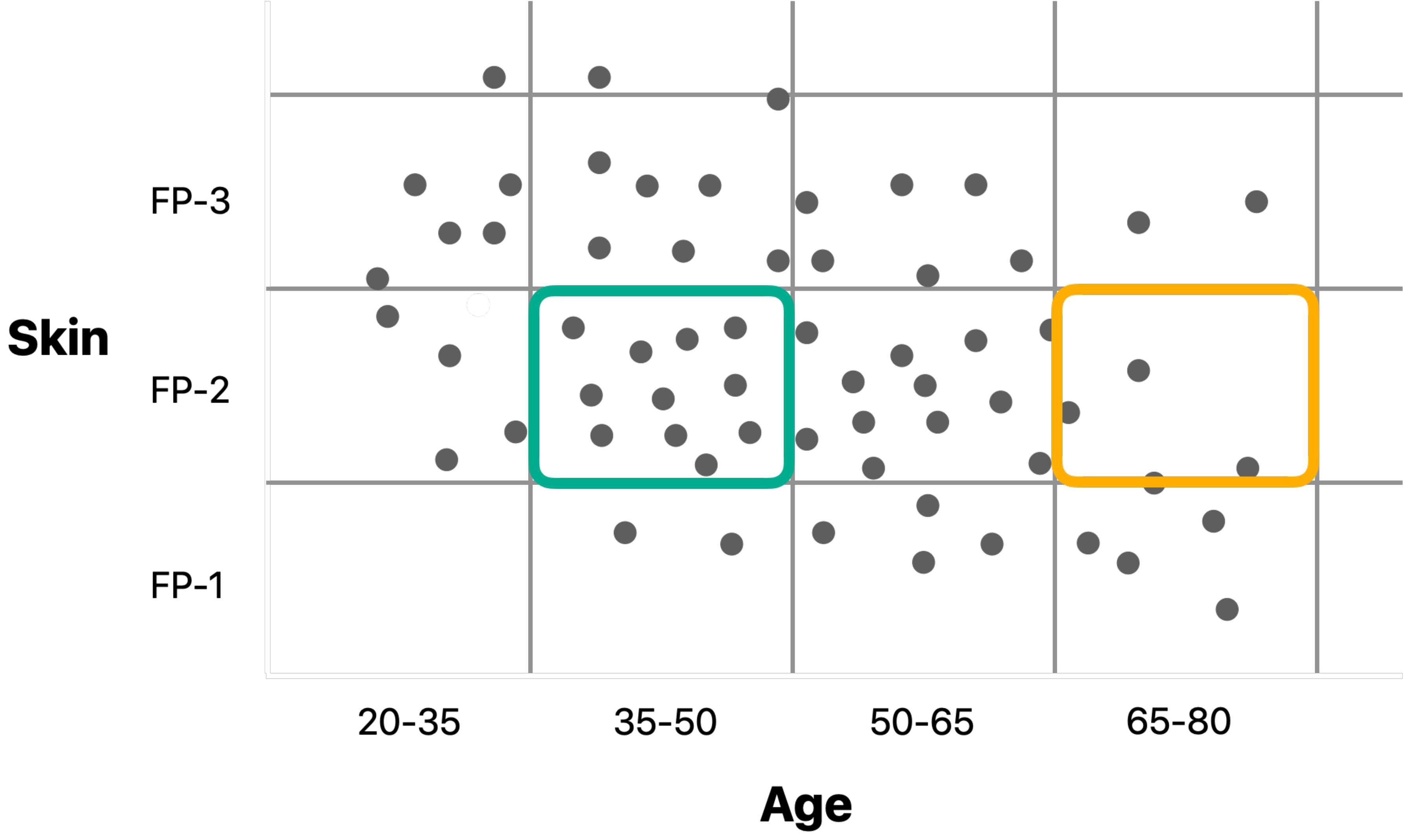}
    }%
    \subfigure[\small{Sharing information with model-based estimates}]{
      \includegraphics[height=1.7in]{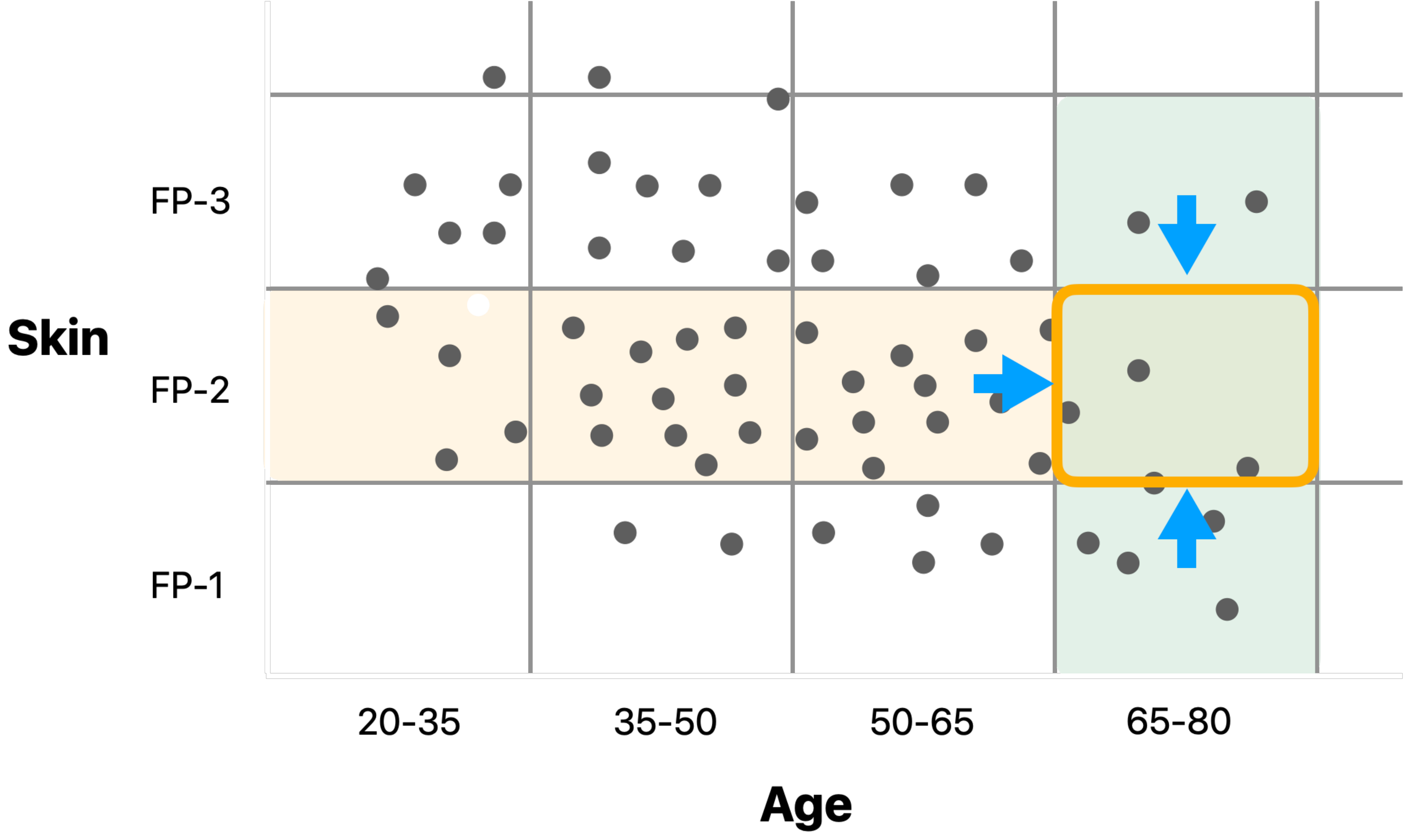}
    }
    \subfigure[\small{Subpopulation frequencies --- NHANES}]{
       \includegraphics[width=.8\textwidth]{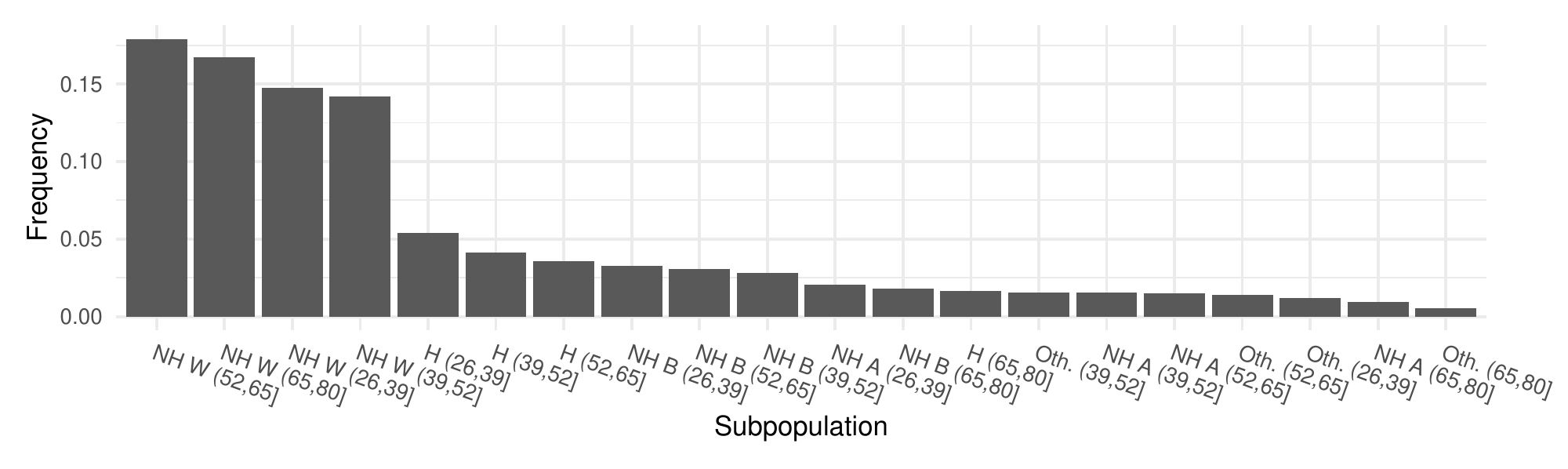}
    }
  }
  \label{fig:overview}
\end{figure*}

Following this idea, we propose the use of a meta-analytic \emph{evaluation model} to estimate the subpopulation performance of a \emph{predictive model} (e.g. the hypothetical cardiac risk model).
The evaluation model approximates the distribution of the predictive model's score conditioned on subpopulation and other covariate information, and is used to form what we will call a \emph{model-based metric} (MBM) estimate.
The evaluation model is fit using all of the data, and by design can incorporate information from the entire sample into each subpopulation estimate. 
Figure~\ref{fig:overview} illustrates this partial pooling idea and highlights subpopulation size imbalances in the US.

We propose a procedure to fit, validate, and ultimately use evaluation models to estimate common performance metrics within subpopulations.
To validate the evaluation models, we use stratified cross validation to compare out-of-sample log-likelihoods between various evaluation models and to a simple nonparametric kernel density estimate as a baseline.
To obtain confidence intervals for the downstream MBM estimates of model performance within subpopulations, we use bootstrap samples.
Furthermore, we propose a novel approximate bootstrap procedure based on importance weighting to avoid fitting a new, computationally expensive evaluation model for each bootstrap replicate (e.g. for Bayesian multilevel models fit via Markov chain Monte Carlo).

In Section~\ref{sec:experiments} we study the empirical performance of our estimators.
In many scenarios of interest where only a small dataset has been collected, ground truth metrics are unavailable, complicating the validation of the proposed estimators. 
As such, we construct two experimental scenarios: (i) a semi-synthetic prediction task that uses demographic and biomarker statistics that match the US population for a cardiovascular risk prediction application, and (ii) a hospital readmission prediction task using data from a large, multi-center study including various clinical attributes \citep{strack2014impact}.
In each setting, we either simulate or use the large dataset to compute ground truth performance metrics in each subpopulation. 
We then compare these ground truth values to estimates derived from smaller subsamples of the full data, mimicking the more typical restricted data setting.
We compare the accuracy and coverage statistics for subpopulation subsample estimates with the proposed MBM estimates. 
Our findings demonstrate that our approach can be much more accurate in smaller subpopulations than the typical subsample estimator.

\subsection*{Generalizable Insights about Machine Learning in the Context of Healthcare}
We show that sophisticated statistical techniques can improve the evaluation of ML model performance in smaller subpopulations. 
Measuring model performance is a crucial part of developing machine learning models to be integrated into clinical settings. 
In health applications, it is especially important to understand the pattern of successes and failures of a model on a diverse set of subpopulations.
Increasing the precision with which we estimate a model's performance on small subgroups can help us build models that are more accurate and fair --- they can highlight subpopulations where we may want to improve the model, or collect additional data through targeted recruitment efforts. 
Through a thorough empirical study, we examine the benefits and shortcomings of our approach, and discuss how we might cope with problems that evade the evaluation model. 
For instance, we highlight when it may be best to revert back to the traditional subsample estimators.

In many health applications we are data-bound, constrained by whatever retrospective data has already been collected.
The aim of this work is to squeeze as much information as possible out of the data at hand. 
More broadly, we adopt the view that predictive model evaluation is itself a data analysis problem, and we should use all of the statistical tools available to help us understand our machine learning models and their potential effects and limitations.

\section{Background}
\label{sec:background}
Consider the setting where a statistical machine learning model (i.e., the \emph{prediction model}) outputs a continuous-valued score to predict a binary class label.\footnote{The extension to multiple discrete categories is straightforward, and we leave the extension to real-valued outcomes to future work.}
For binary classification, the predictive model score is simply the log conditional probability of the positive class, $\ln Pr(Y = 1 \,|\, \texttt{input})$, where the data $\texttt{input}$ is whatever input data the predictive model conditions on (e.g., an image, laboratory values, vital signs).
Additionally, we observe subpopulation information and additional relevant covariates.
We denote these variables:
\vspace{-.5em}
\begin{itemize} \itemsep 0pt
    \item $Y \in \mathcal{Y} = \{0, 1\}$: the binary outcome of interest, e.g., low or elevated disease risk; 
    \item $S \in \mathbb{R}$: the predictive model score, e.g., from a machine learning model.  In a binary classification setting, we define $S \triangleq \ln Pr(Y=1 \,|\, \texttt{input})$;  
    \item $A \in \mathcal{A}$: the discrete subpopulations defined by demographic attributes, e.g., age, sex, and race categories;
    \item $X \in \mathcal{X}$: additional covariates to condition on, e.g., BMI or cholesterol.
\end{itemize}
\vspace{-.5em}
Our goal is to measure the prediction model's performance on each subpopulation as precisely as possible. 
Given a predictive model with score $S=s$ and true class $Y=y$, we evaluate the model using a set of performance metrics.\footnote{We use lower case letters to denote realizations of these random variables (e.g., $A = a$).}  

As a concrete example, consider the \emph{area under the receiver operating characteristic curve} (AUC) --- a ubiquitous performance metric --- conditioned on a particular subpopulation $a$. 
The AUC is equivalent to the probability of correctly ranking two independent observations, $i$ and $j$, one from each class:
\begin{align}
    AUC(a) &= Pr(S_j > S_i \,|\, A_i = A_j = a, Y_i = 0, Y_j = 1) \\
           &= \mathbb{E}_{Pr(S \,|\, Y=1, A=a) Pr(S' \,|\, Y=0, A=a)}[\boldsymbol{1}(S > S')] \, .
\end{align}
A high AUC indicates that the model prediction is more likely to correctly rank two randomly selected individuals with different outcome ($Y$) values, while an AUC of $0.5$ indicates that a model performs this ranking no better than chance.
Crucially, note that the AUC is a function of the \emph{class conditional distributions} of the prediction model score, $Pr(S \,|\, Y=1, A=a)$ and $Pr(S' \,|\, Y=0, A=a)$.

Another common metric is the \emph{positive predicted value} (PPV) at threshold $\tau$ (i.e., when $S > \tau$, the model predicts a positive label $Y=1$).  Again, this metric can be expressed as a function of the class conditional distributions of the prediction model score,
\begin{align}
    PPV_{\tau}(a) &= \frac{Pr(S>\tau, Y=1 \,|\, A=a)}{Pr(S>\tau, Y=1 \,|\, A=a) + Pr(S>\tau, Y=0 \,|\, A=a)},
\end{align}
which is a function of both the class conditional distributions of the score $S$ and the overall prevalence of positive examples within subpopulation $a$. PPV offers a different view into model performance, as it is the probability that a subject is actually in the positive class $Y=1$ given a positive prediction $S > \tau$. 

Lastly, the \emph{false positive rate} (FPR) describes the frequency of negative examples incorrectly classified to be positive, and can be written
\begin{align}
    FPR_{\tau}(a) &= Pr(S > \tau \,|\, Y=0, A=a) = \mathbb{E}_{Pr(S \,|\, Y=0, A=a)} \left[ \boldsymbol{1}(S > \tau) \right] \,,
\end{align}
which is a function of only the $Y=0$ class conditional distribution. 

In general, we denote the subpopulation performance metric we wish to estimate as $\theta(a)$, a functional of the class conditional distributions,
\begin{align} 
  \theta(a) &= f\Big( \big\{Pr(S \,|\, Y=y, A=a)\big\}_{y \in \{0,1\}} \Big) \,, 
\end{align}
where the form of $f(\cdot)$ specifies the desired performance metric. 

\subsection{Non-parametric AUC, PPV, and FPR estimators}
Typically we only have access to a sample from these conditional distributions, necessitating sample-based estimators of these metrics. 
Denote the observed data $\mathcal{D} = \{ (a_n, x_n, y_n, s_n) \}_{n=1}^{N}$, corresponding to, respectively, demographic attributes, covariates, the true outcome class, and the ML model score for subject $n$ for each of $N$ total subjects.  
Let $N_y = \{n : y_n = y\}$.

\vspace{-.5em}
\paragraph{AUC estimators: The normalized Mann-Whitney U-Statistic.} 
A common estimator for the AUC is the unbiased Mann-Whitney U-statistic.
Given a sample of scores from positive ($y_n=1$) examples $\{s_n : n \in N_1 \}$ and negative ($y_n=0$) examples $\{s_n : n \in N_0\}$, the U-statistic estimator of the AUC is 
\begin{align}
    \hat{\theta} &= \frac{1}{|N_1||N_0|} \sum_{n_1 \in N_1}\sum_{n_0 \in N_0} \boldsymbol{1}(s_{n_1} > s_{n_0}) \, .
\end{align}

To estimate frequentist confidence intervals, a common strategy is to compute statistics of bootstrap samples \citep{efron1992bootstrap}.
Another common way to compute the AUC metric is by numerically integrating the empirical ROC curve directly. This can be accomplished in quasi-linear time and may be more suitable for large samples. 

\vspace{-.5em}
\paragraph{Threshold estimators: FPR, PPV}
To compute metrics with respect to a fixed threshold $\tau$, estimators are typically simple functions of the confusion matrix.  
For example, the false positive rate is typically estimated by the empirical frequency of negative examples incorrectly classified as positive,
\begin{align}
    FPR_\tau &= \frac{1}{N} \sum_{n_0 \in N_0} \boldsymbol{1}(s_{n_0} > \tau) \,  .
\end{align}

Likewise, the positive predictive value is the empirical proportion of true positives among the set of samples classified as positive,
\begin{align}
    PPV_\tau &= \frac{|\{n : s_n > \tau, y_n=1 \}| }{ |\{ n : s_n > \tau \}|} \, .
\end{align}

To form a subpopulation-specific estimate for group $A=a$, all of these non-parametric sample-based estimators are simply restricted to examples from $a$.
This restriction reduces the sample size, increasing the variance of the estimator. 
For large subpopulations, the scale of the variance may be small enough to be acceptable, but for smaller subpopulations the high variance caused by this restriction can result in uninformative estimates of performance.


\section{Model-based metric evaluation}
The non-parametric sample-based estimators described in the previous section are simple to implement and admit favorable theoretical properties (e.g., unbiasedness).  However, naively applying these estimators to small subpopulations will yield noisy estimates.
We propose \emph{model-based metric} (MBM) estimates to form sample-efficient estimates of common predictive model metrics.  The key idea is that when forming an estimate for subpopulation $a$, we can borrow information from other, similar subpopulations. 
To borrow information, we specify an \emph{evaluation model} --- a joint model of the prediction score $S$ given class $Y$, subpopulation $A$, and covariate information $X$ --- over the entire observed population.  Specifying a model also allows us to incorporate information from covariates, which can help us more precisely specify this joint distribution. 
As discussed in the previous section, common performance metrics are functions of these conditional distributions, and can be computed using the fitted evaluation model.

Using an evaluation model is similar to specifying any model to analyze a dataset, requiring a plausible parametric form, an algorithm to fit the model to observations, and a procedure to criticise and choose between models.  Here, we describe the steps of our model-based metric evaluation process: (i) specifying the evaluation model, (ii) fitting the evaluation model to data, (iii) checking and validating model fit, (iv) computing model-based metric estimates, and (v) efficiently computing confidence intervals for these model-based estimates. 

\subsection{Evaluation model specification}
The evaluation model is a parametric model of the class conditional distribution of the model score, parameterized by $\lambda$, which we denote
\begin{align}
    S \,|\, \{ A=a, X=x, Y=y \} &\sim Pr_\mathcal{M}(S \,|\, A=a, X=x, Y=y, \lambda) \, ,
\end{align}
where $\mathcal{M}$ indicates that this conditional distribution is defined by the chosen model class $\mathcal{M}$, $a$ indicates the subpopulation, $x$ are additional covariates, and $y$ is the true (binary) class. 
As a concrete example, for a continuous-valued $S$, a common assumption might be that the response is conditionally Gaussian, 
\begin{align}
    S \,|\, \{ A = a, X = x, Y = y \} &\sim \mathcal{N}(\mu_{a, x, y}, \sigma_{a, x, y}^2) \, ,
\end{align}
where the mean and variance of this model, $\lambda = \{\mu_{a,x,y}, \sigma^2_{a, x, y} : a \in \mathcal{A}, x \in \mathcal{X}, y \in \mathcal{Y} \}$, are determined by the subpopulation, covariate, and class information.

If we treat each subpopulation determined by $a, x$, and $y$ independently (``no-pooling,'' in the parlance of multi-level modeling), we recover a parametric version of the fully independent subsample estimators described in Section~\ref{sec:background}.
However, when we specify shared structure in the parameters $\mu_{a,x,y}$ and $\sigma^2_{a,x,y}$, the model-based subpopulation estimates can use relevant information from related subpopulations --- in essence exploring the bias-variance tradeoff. 
In our empirical analysis, we explore a range of linear models, from simple additive models with homoscedastic errors to more complex multi-level models with pairwise effects and heteroscedastic errors (see Appendix~\ref{sec:app-evaluation-models} for details). 
We emphasize that our approach is fully general, and that many other functional forms for evaluation models are possible for other applications.

\subsection{Evaluation model inference}
For a given evaluation model class $\mathcal{M}$, we must fit the evaluation model parameters $\lambda$ to best describe the observed predictive model score data using the available demographic and covariate information. Adopting a Bayesian approach, we integrate over our posterior uncertainty in $\lambda$ to form the class conditional distributions of interest,
\begin{align}
    \label{eq:posterior-predictive}
    Pr_{\mathcal{M}}(S \,|\, A=a, X=x, Y=y, \mathcal{D}) 
      &= \int Pr_{\mathcal{M}}(S \,|\, A=a, X=x, Y=y, \lambda) Pr_{\mathcal{M}}(\lambda \,|\, \mathcal{D}) d\lambda \, .
\end{align}
This is the posterior predictive distribution, which is the model likelihood averaged over the posterior $Pr_{\mathcal{M}}(\lambda \,|\, \mathcal{D})$ given \emph{all of the data} $\mathcal{D}$. This averaging results in a more expressive mixture distribution than the model conditioned on a single value of $\lambda$.

As discussed above, metrics of interest --- such as AUC, PPV, and FPR --- are functions of this class conditional distribution. Analogously, our model-based metric estimates are functions of the posterior predictive distribution of Equation~\ref{eq:posterior-predictive}. To compute such metrics, we use Markov chain Monte Carlo (MCMC) to approximate the posterior predictive distribution.  We simulate $R$ samples, $\{ \lambda^{(r)} \}_{r=1}^R$ from the posterior distribution using the No U-turn Sampler (NUTS) \citep{hoffman2014no}, a variant of Hamiltonian Monte Carlo. 
These samples can then first be used to perform model checking and validation, and then compute model-based metrics as described in Section~\ref{sec:method-mbm}. 

\subsection{Evaluation model checking}
When forming a model-based estimate, a first order concern is the fidelity of the evaluation model to the true distribution generating the observed prediction model scores.  We suggest two validation procedures: (i) a set of posterior predictive checks, and (ii) cross validated log-likelihood comparisons.
We compare the out-of-sample log-likelihood estimates between evaluation models, and also compare against a kernel density estimate (KDE) formed within each subpopulation $A=a$ as a baseline.
The KDE is a surrogate model for the empirical metrics --- if the KDE's out-of-sample log likelihood is significantly better than the evaluation model's, this indicates that the evaluation model is likely overfit and will yield invalid results.  
In this event, we iterate on the model class.  
If no evaluation model is successful in comparison to the KDE, we suggest reverting back to the non-parametric subsample estimator. 

\subsection{Computing model-based metric estimates} 
\label{sec:method-mbm}
To compute model-based metric estimates, we use samples from the posterior predictive distribution.
For each posterior sample $\lambda^{(r)}$, the model scores $s_1, \dots, s_n$ are simulated from the model conditional distribution, resulting in simulations $\{s^{(sim)}_{n,r}, y_{n}, a_n, x_n\}_{n=1,r=1}^{N,R}$.
Intuitively, this is a larger dataset of $N\times R$ model-based simulations that can then be plugged into the standard nonparametric AUC, FPR, and PPV estimators, resulting in an approximation of the Bayes estimate of each model metric. 

\subsection{Confidence intervals and approximating the bootstrap}
\label{sec:approximate-bootstrap}
For an evaluation model $\mathcal{M}$ and subpopulation, the theoretical Bayes estimate of a model metric is a \emph{point estimate} --- a scalar that is a deterministic function of the observed data $\mathcal{D}$.
To see this, consider that as the number of posterior simulations $R$ grows, we eliminate Monte Carlo error from the MCMC sampling routine.
As such, the posterior predictive class conditional distributions become \emph{deterministic} functions of $a$, $x$, $y$, and $\mathcal{D}$, making the corresponding metric estimate also deterministic (though in practice, with finite $R$, there will be a small amount of Monte Carlo error).

Typically, we also want to characterize the uncertainty of this point estimate, for example, by computing frequentist confidence intervals.
One popular approach is the bootstrap \citep{efron1992bootstrap}. 
In our setting, this involves sampling a boostrapped dataset $\mathcal{D}_b$, simulating from the posterior predictive distribution given $\mathcal{D}_b$, and computing the model-based metric estimate --- repeated $B$ times, where practically $B$ is at least 100. 
Bootstrapping Bayes estimates is similar to using a bagged posterior \citep{buhlmann2014discussion, huggins2019robust, huggins2020robust}. 

While conceptually straightforward, it can be computationally prohibitive to simulate $R$ posterior samples $B>100$ times, particularly for hierarchical models with high-dimensional parameters $\lambda$. 
For such models, we propose an approximation to the bootstrap that reuses a single set of posterior samples given the full data $\mathcal{D}$. 
To generate a single bootstrap estimate, we re-weight each posterior sample in a way that approximates the bootstrapped posterior.
This approach follows a similar strategy to the approximate leave-one-out cross validation estimates developed in \citet{vehtari2017practical} and \citet{vehtari2015pareto}.

Concretely, given a set of posterior simulations, $\Lambda \triangleq \{ \lambda^{(r)} \}_{r=1}^R$, $\lambda^{(r)} \sim Pr_{\mathcal{M}}(\lambda \,|\, \mathcal{D})$, we use importance-weighted posteriors to approximate a set of bootstrapped estimators. 
For each bootstrap dataset $\{\mathcal{D}_b\}_{b=1}^B$, we compute truncated self-normalized weights for each posterior sample $r$ as follows:
\begin{align}
    \tilde{w}_{b,r} 
      &= \frac{Pr_{\mathcal{M}} \left( \lambda^{(r)} \,|\, \mathcal{D}_b \right) }{Pr_{\mathcal{M}} \left( \lambda^{(r)} \,|\, \mathcal{D} \right) } 
      = \frac{Pr_{\mathcal{M}} \left( \mathcal{D}_b \,|\, \lambda^{(r)} \right) }{Pr_{\mathcal{M}} \left( \mathcal{D} \,|\, \lambda^{(r)} \right) } \\
    \tilde{w}^{(trunc)}_{b,r} &= \min \left( \tilde{w}_{b,r}, \sqrt{R} \cdot \frac{1}{R} \sum_{r=1}^R \tilde{w}_{b,r} \right) \\
    w_{b,r} &= \frac{\tilde{w}^{(trunc)}_{b,r}}{\sum_{r=1}^R \tilde{w}^{(trunc)}_{b,r}}   \, .
\end{align}
The evaluation model prior terms cancel out, simplifying to the likelihood ratio between the bootstrap dataset $\mathcal{D}_b$ and the original dataset $\mathcal{D}$. 
Next, we construct a set of posterior samples approximately distributed according to the \emph{bootstrapped posterior}, $\lambda_b^{(r)} \sim Pr_{\mathcal{M}}(\lambda \,|\, \mathcal{D}_b)$, by sampling from the original set of posterior samples $\Lambda$ proportional to the truncated weights with replacement, $\lambda_b^{(r)} \sim \text{Categorical}(\Lambda, w_{b})$, resulting in an approximate bootstrap posterior sample $\Lambda_b = \{ \lambda_b^{(r)} \}_{r=1}^R$
We then compute the desired model-based metrics for each sample $\Lambda_b$, and repeat the process $B$ times.

Crucially, the importance-weighted approximation requires computing posterior samples for only one model, while the typical bootstrap estimator requires computing posterior samples for \emph{each} of the $B$ bootstrapped datasets.
Figure~\ref{fig:ci-comparison} in the Appendix shows that the confidence intervals for different performance metrics obtained via this efficient approximation closely resemble the intervals obtained from the standard (expensive) bootstrapping procedure.

\section{Method summary}
To summarize, we propose the following procedure to produce model-based estimates of common ML model performance metrics in subpopulations:
\vspace{-.5em}
\begin{itemize} \itemsep 0pt
    \item Specify an evaluation model, $Pr_{\mathcal{M}}(S \,|\, A=a, X=x, Y=y, \lambda)$.
    \item Simulate posterior samples from the evaluation model, $\{ \lambda^{(r)} \}_{r=1}^R$, $\lambda^{(r)} \sim Pr_{\mathcal{M}}(\lambda \,|\, \mathcal{D})$, given a dataset $\mathcal{D}$ of predicted scores from the ML model.
    \item Validate the evaluation model by estimating out-of-sample performance with stratified cross validation (or approximate leave-one-out cross validation \citep{vehtari2015pareto})
    \vspace{-.5em}
    \begin{itemize} \itemsep 0pt
        \item Compare to subpopulation-specific KDE log-likelihoods: if the evaluation model underperforms the KDE on a subpopulation, revert to the typical subsample-based estimate of model performance, or revise the evaluation model and refit.
    \end{itemize}
    \vspace{-.5em}
    \item Form model-based metric estimates for each subpopulation from posterior predictive simulations $\{ s^{(sim)}_{n,r} \}$, $\theta(x) = f(\{ s^{(sim)}_{n,r} \})$. 
    \item Form bootstrap-simulated model-based metric estimates:
    \vspace{-.5em}
    \begin{itemize} \itemsep 0pt
        \item Simulate bootstrap dataset $\mathcal{D}^*_b$.
        \item Compute posterior samples using $\mathcal{D}^*_b$ (alternatively, approximate $Pr_{\mathcal{M}}(\lambda \,|\, \mathcal{D}^*_b)$ with importance weights).  
        \item Compute relevant model-based metric estimates $\theta^*_b(x)$ for each subpopulation $x$.
    \end{itemize}
    \vspace{-.5em}
    \item Return point estimate $\theta(x)$ and bootstrap samples $\{ \theta^*_b(x) \}_{b=1}^B$.
\end{itemize}
\vspace{-.5em}
This procedure results in a point estimate and bootstrap samples of the performance metrics of interest. In addition to the subpopulation-specific metrics, comparisons between subpopulations can also be computed using the evaluation model posterior samples. See Appendix~\ref{sec:app-evaluation-models} for evaluation models used in our empirical study. 

\subsection{Related work}

The introduction of a model and Bayesian inference for estimating model metrics has been discussed before, for computing accuracy \citep{benavoli2017time}, true positive rate \citep{johnson2019gold}, and precision-recall \citep{goutte2005probabilistic}. 
The most similar work to ours focuses on the use of unlabeled data to tighten estimates of subpopulation performance, using a model parameterized directly by the metric of interest (e.g., TPR) \citep{ji2020can}.  
This work differs from ours in two important ways: (i) their approach focuses on specific threshold-based metrics (e.g., TPR, FPR), and does not address statistics like the AUC or other concordance metrics, and (ii) uncertainty is characterized by Bayesian credible intervals, and not frequentist confidence intervals. Our framework is more general, as we model the class conditional distribution of the predictive model score directly, from which all metrics of interest can be computed in a single procedure.  
As such, our evaluation models are not restricted to just one metric defined at one particular threshold.

Regarding estimate uncertainty, in a series of articles, \citet{huggins2019robust} and \citet{huggins2020robust} discuss ``bagged posteriors'', or an ensemble of posterior distributions conditioned on bootstrapped datasets for forming robust estimates and for model selection.  They explore the theoretical properties of such estimators, showcasing their robustness in the presence of model misspecification, motivating their use in constructing our estimator.
They note that the bagged posterior combines the desirable properties of Bayesian and frequentist methods: flexible hierarchical estimators that average out nuisance parameters with robustness to sampling variability. 

Our problem setting and model-based approach are similar in spirit to the goals of meta-analysis \citep{lipsey2001practical}, where Bayesian multi-level models have been used extensively to partially pool information across related groups and studies \citep{gelman2013bayesian}.
\citet{oakden2020docs} discusses the intersection of meta-analysis and machine learning model evaluation, focusing on the comparison of model predictions to predictions from a panel of multiple experts.  This work makes clear that simple averages of sensitivity and specificity across experts typically underestimate expert performance.  Combining expert labels requires a meta-analytic approach, which is conceptually related to the way we construct model-based estimates of metrics such as AUC.

Lastly, \citet{hanczar2010small} discuss reliability of small-sample ROC (and FPR) estimates in biological settings. 
They caution that a procedure for error estimation can only provide as much information as is present in the data at hand; unrepresentative samples will yield poor error estimates.  We observe the same phenomenon, and we attempt to address such shortcomings within individual subpopulations by incorporating related information into our subpopulation estimates.

\section{Empirical evaluation}
\label{sec:experiments}
We apply model-based metric estimates in two settings. First, we consider a semi-synthetic scenario using demographic and biomarker statistics from the National Health and Nutrition Examination Survey (NHANES) \citep{nhanes} to mimic a cardiovascular disease risk prediction task, creating a scenario where ground truth metrics are available. Then, we evaluate a model for predicting hospital readmission among diabetes patients, using data from a large, multi-center study \citep{strack2014impact}.
To specify and fit evaluation models, we use \emph{Stan} \citep{carpenter2017stan} and the \emph{brms} package in R \citep{brms}.

\subsection{Semi-synthetic Framingham risk prediction (\texttt{semi})}
\vspace{-.5em}

\paragraph{Problem setup}
We set up a semi-synthetic prediction task that targets the Framingham risk score, a metric of cardiovascular health that aggregates age, blood pressure, cholesterol, diabetes, hypertension, and smoking status \citep{d2008general}.  
To simulate an observation, we first randomly draw from groups defined by age bins, sex, race, and BMI in proportion to their empirical frequencies in NHANES.  
For each simulated individual, we then sample Framingham risk factors from group-specific aggregated statistics in the NHANES population.
We construct an outcome variable by computing the Framingham risk score \citep{d2008general}, and using 10\% risk as a cutoff to define two classes --- low and elevated risk of cardiovascular disease \citep{bosomworth2011practical}.\footnote{We emphasize that this is a hypothetical prediction task, and highlight that the development of the Framingham cardiovascular risk score itself suffered from exactly the subpopulation recruitment issues that we describe, necessitating follow-up studies targeting underrepresented groups \citep{kanaya2013mediators}.
}

To mimic a machine learning model predicting this synthetic target, we simulate prediction model scores $S$ in three ways: (i) a score with no relationship to subpopulation group and covariates (\texttt{none} in the results), (ii) a score with a simple additive relationship to age, race, sex, and BMI (\texttt{simple}), and (iii) a score with interactions between race and age, and sex and BMI (\texttt{interactions}). 
Additionally, for each type of simulated prediction model scores, we add either homoscedastic or heteroscedastic (denoted \texttt{-hetero} in the results) noise.
See Appendix~\ref{sec:app-nhanes} for full details of each of the six settings.

For each of these six simulated ML models, we generate a large dataset with $N^{(pop)}=5 \times 10^6$ observations to use as ground truth.  
We then take a smaller subsample (e.g., $N=10{,}000$, $N=5{,}000$, and $N=1{,}000$) to simulate a more realistic limited data setting.  
We construct MBM estimates and the nonparametric subsample-based estimates using this smaller dataset.  

\vspace{-.5em}
\paragraph{Evaluation models}
We report the AUC, FPR, and PPV metric estimates from both the typical subsample estimator (denoted \texttt{empirical}) and a set of evaluation models --- thresholds for FPR and PPV are set such that the full population FPR is 1\%. 
We use a set of evaluation models of varying complexity:
\vspace{-.6em}
\begin{itemize} \itemsep 0pt
    \item \texttt{fixed.b}: a fixed effects model with the following linear predictors: demographics (gender, race, age bin), BMI, true class $Y$, and  additional covariates (blood pressure and lipids).
    \item \texttt{fixed.c}: a fixed effects model containing all marginal and pairwise interactions between demographics, BMI, and $Y$, as well as the additional covariates in \texttt{fixed.b}.
    \item \texttt{rand.a}: a random effects model consisting of a random intercept for all marginal and pairwise interactions between demographics, BMI, and $Y$, as well as the additional covariates in \texttt{fixed.b} as fixed effects.
\end{itemize}
\vspace{-.6em}
We also explored heteroscedastic evaluation models with a noise variance modeled as log-linear with subpopulation covariates, but found results to be similar to those with constant variance.
Full experiment and evaluation model details are in Appendix~\ref{sec:app-nhanes}.

\subsection{Diabetes hospital readmission prediction (\texttt{dm})} 

\vspace{-.5em}
\paragraph{Problem setup}
We also apply MBMs to a more complex scenario involving a real healthcare prediction task.
We use a large, publicly available health dataset that was constructed to understand potential factors driving hospital readmissions among individuals with diabetes \citep{strack2014impact, Dua:2019}. 
After filtering to only adult admissions and removing observations with missing demographics, the dataset consists of $96{,}229$ hospital admissions among $67{,}467$ unique individuals.
The prediction problem is to use information available on discharge (e.g. length of stay, principal diagnoses, hemoglobin A1c result, diabetes medications) to predict the binary outcome whether the individual is ever readmitted to the hospital for a diabetes-related problem. 

We first construct an XGBoost \citep{Chen:2016:XST:2939672.2939785} model to predict the binary outcome of interest, using a total of 105 features after categorical variables are one-hot encoded, treating the predictions made by this model as a form of ground truth.
We use this predictive score to compute subpopulation AUCs (and FPRs and PPVs) using the typical non-parametric estimators.
Next, we take a random subsample of size 5{,}000 patients, and apply our MBM estimators to the subsample.
We then draw similar comparisons as for \texttt{semi} --- we measure the accuracy and coverage of model-based estimators and the subsample non-parametric estimators to the full-sample non-parametric estimators.  
We repeat this sub-sampling procedure 10 times to estimate statistics over the data distribution. 

\vspace{-.5em}
\paragraph{Evaluation models}
We use a similar set of evaluation models as in the \texttt{semi} task, differing only in the specific subpopulations and additional covariates.
\texttt{fixed.b} again denotes a fixed effects model with demographic subpopulation and additional covariates as linear predictors.
Similarly, \texttt{fixed.c} contains the same variables as in \texttt{fixed.b}, but now allows for all marginal and pairwise interactions between demographic variables.
Lastly, \texttt{rand.a} is a random effects model with intercepts for all marginal and pairwise interactions between demographic variables, with additional covariates as fixed effects.
Full details on the \texttt{dm} experiment settings and evaluation models can be found in the Appendix~\ref{sec:app-evaluation-models}.

\subsection{Results}

\begin{figure*}[t!]
\floatconts
  {fig:nhanes-experiment}
  {
  \vspace{-1.5em}
  \caption{\small{Results from a run of the \texttt{semi} experiment. 
  Descriptions of each method can be found in the text or Appendix. 
  Results are obtained using a random sample of $10{,}000$ individuals from the total simulated population of five million.
  Abbreviations for demographic subgroups: ``H'': Hispanic, ``NH A'': non-Hispanic Asian, ``NH B'': non-Hispanic African American, ``NH W'': non-Hispanic Caucasian.
  }}
  }
  {%
    \subfigure[\small{AUC estimates and 95\% confidence intervals by subpopulation, under the most complex \texttt{interactions-hetero} simulation.}]{
      \label{fig:nhanes-auc-by-subpop}
      \includegraphics[width=\textwidth]{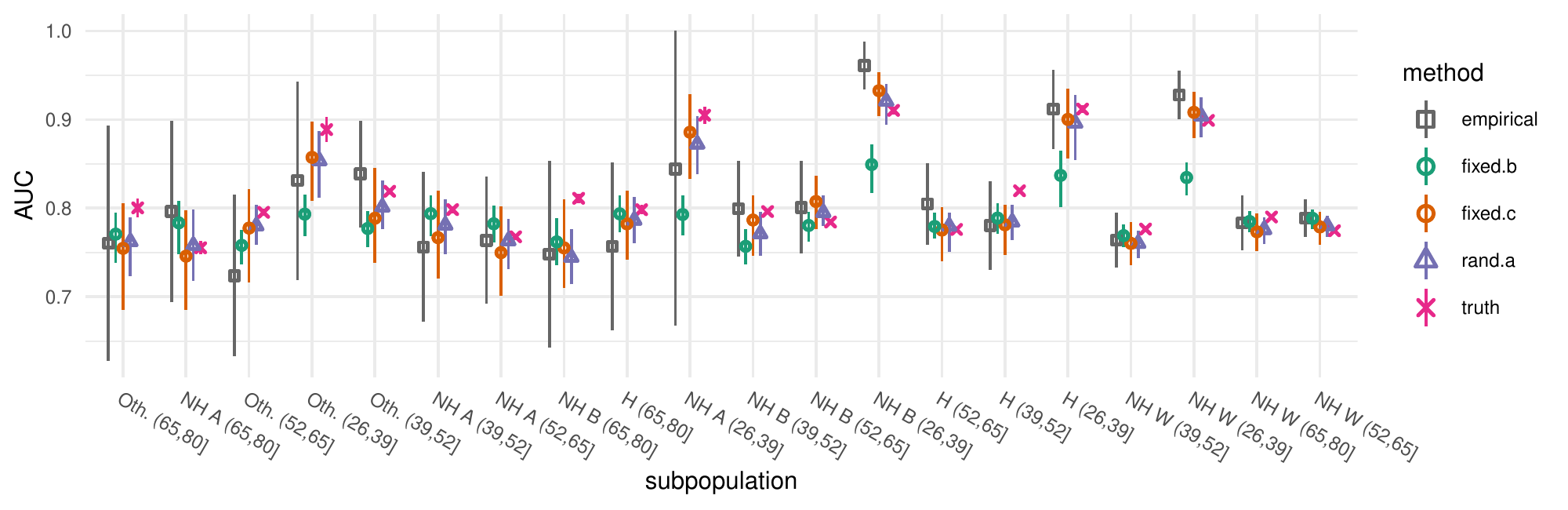}
    }
    \subfigure[\small{Relative mean absolute percentage error (lower is better --- $1$ is scaled to the empirical subsample estimator error).}]{
      \label{fig:nhanes-auc-mape}
      \includegraphics[width=.95\textwidth]{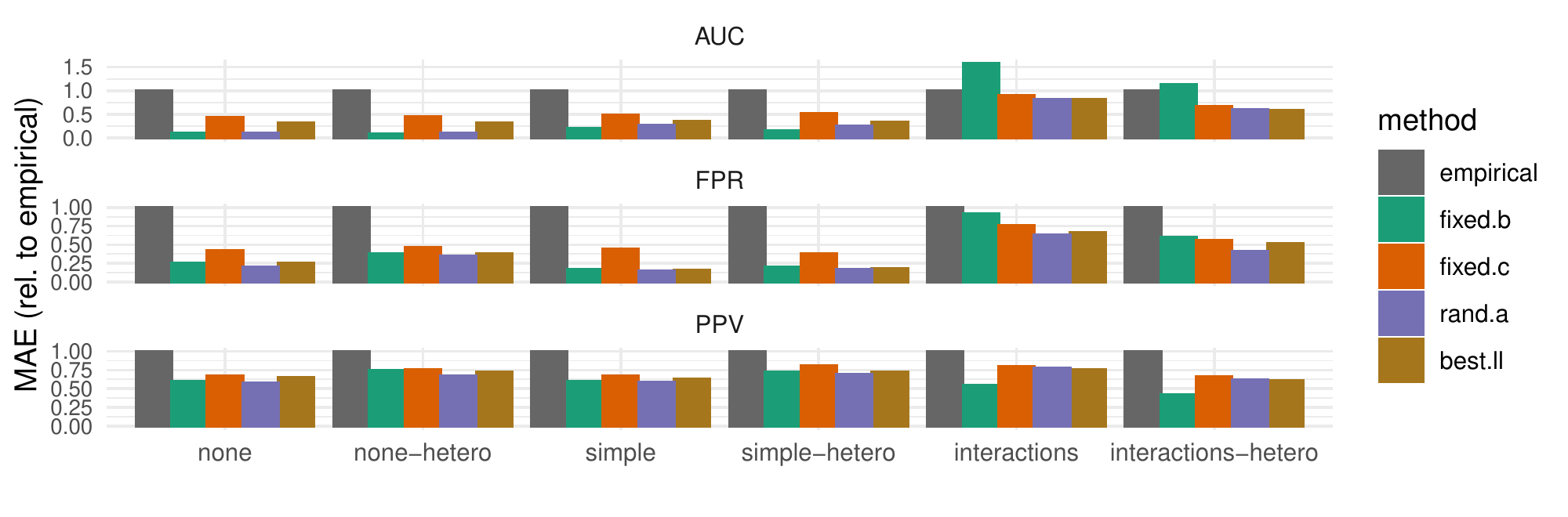}
    }
    \subfigure[\small{Relative negative log-likelihood (lower is better --- $1$ is scaled to the KDE). }]{
      \label{fig:nhanes-nlls}
      \includegraphics[width=.95\textwidth]{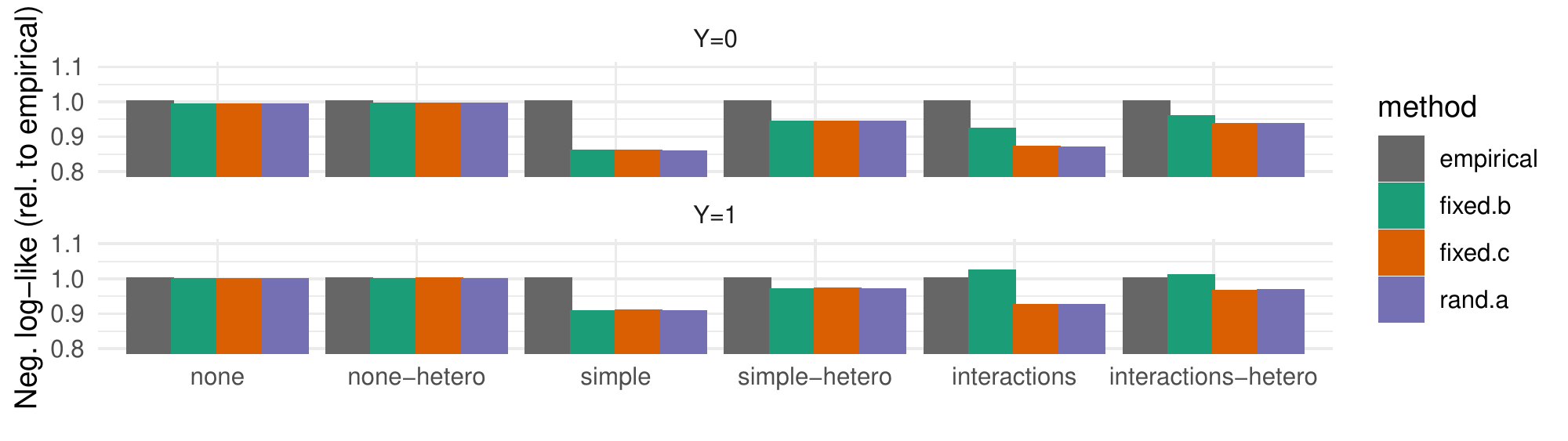}
    }
  }
  \label{fig:nhanes-experiments}
\end{figure*}

\begin{figure*}[t]
\floatconts
  {fig:rehosp-experiment}
  {
  \vspace{-1.5em}
  \caption{\small{Results of the \texttt{dm} experiment. (a) Estimator error as a function of data size --- we observe larger benefits in the smaller sample range across all three statistics.  (b) A breakdown of AUC estimates for each subpopulation, comparing to the ``true'' AUC (based on the full dataset of $96{,}229$ observations). The empirical and various MBM estimates are from subsamples of size 5{,}000. Subpopulation abbreviations: ``H'': Hispanic, ``Oth.'': Other, ``AA'': African American, ``C'': Caucasian, ``M'': Male, ``F'': Female}}
  }
  {%
    \subfigure[\small{Estimator error by sample size}]{
      \label{fig:rehosp-auc-mape}
      \includegraphics[width=\textwidth]{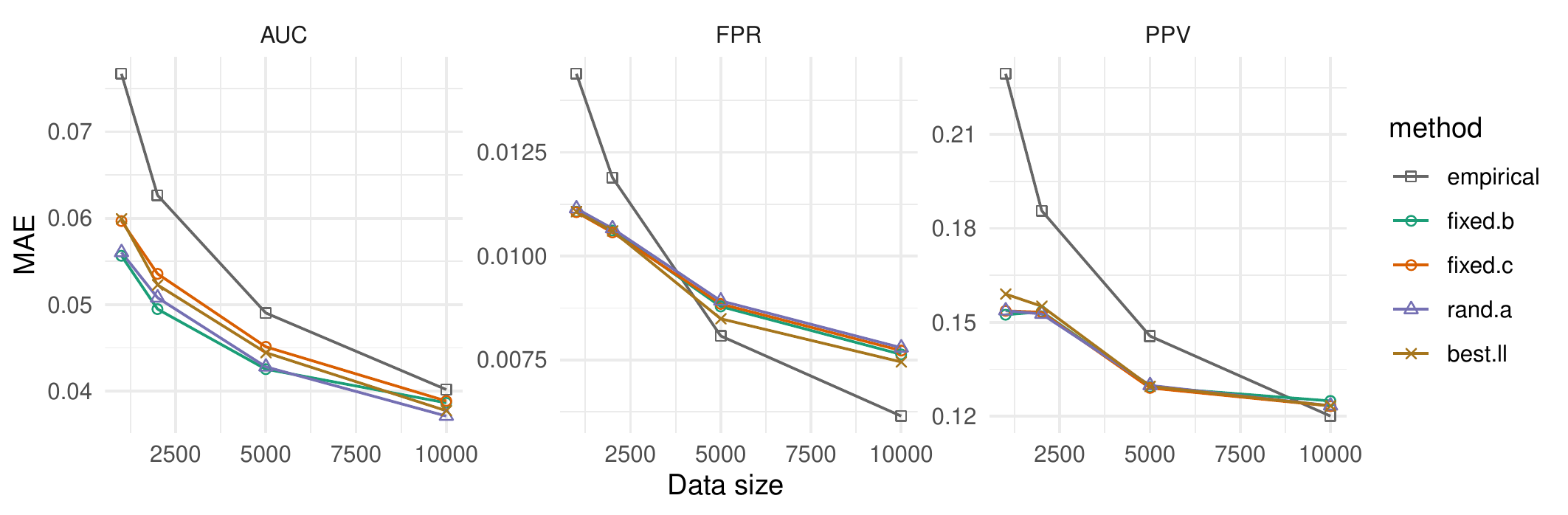}
    }
    \subfigure[\small{Subpopulation AUC estimator comparison}]{
      \label{fig:rehosp-auc-by-subpop}
      \includegraphics[width=\textwidth]{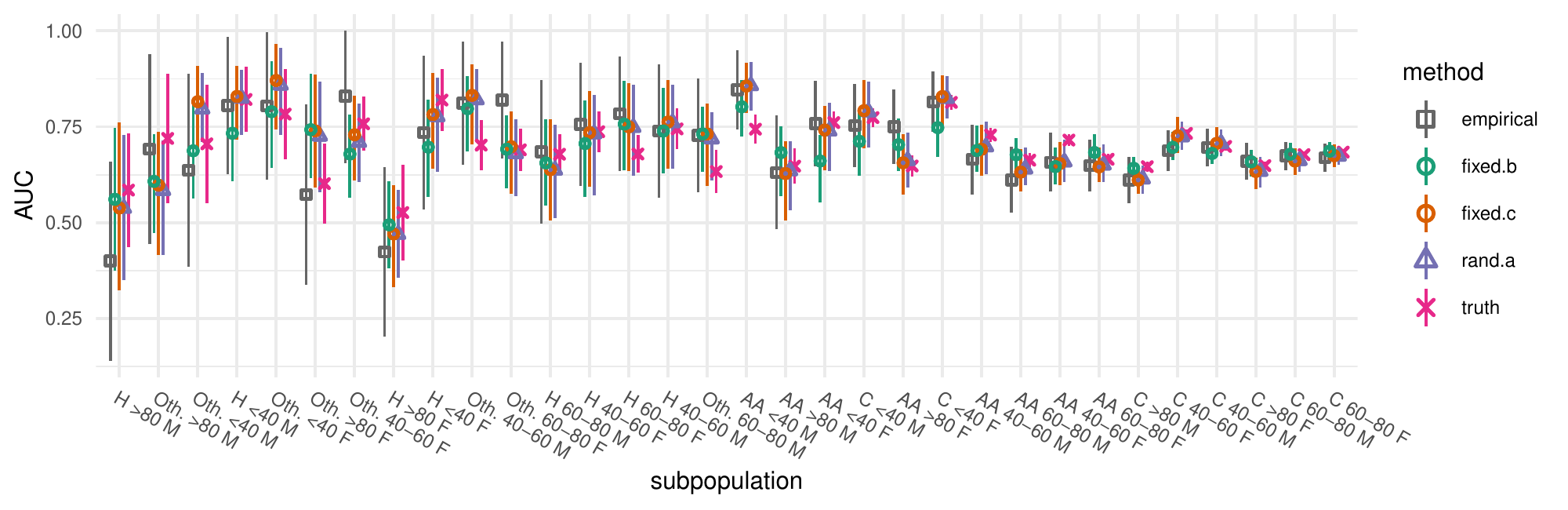}
    }
  }
  \label{fig:rehosp-experiments}
\end{figure*}

\vspace{-.5em}
\paragraph{MBMs generally improve estimates of AUC, FPR, and PPV.}
\Cref{fig:nhanes-experiments} depicts results from the \texttt{semi} task, obtained from a random subset of $N=10{,}000$ observations.
\Cref{fig:nhanes-auc-mape} shows the relative error in estimating AUC, FPR, and PPV, averaged over all subpopulations.
\texttt{best.ll} denotes the results obtained by selecting the best-fitting evaluation model (i.e., either the \texttt{empirical} KDE, or one of \texttt{fixed.b}, \texttt{fixed.c}, or \texttt{rand.a}) within each subpopulation, and then averaging these results across subpopulations.
Across all six simulation settings, at least one of the MBMs produces more accurate estimates of AUC, FPR, and PPV compared to the non-parametric subsample estimates (\texttt{empirical}). 

\vspace{-.5em}
\paragraph{Improvements are most substantial for smaller subpopulations.}
\Cref{fig:nhanes-auc-by-subpop} depicts AUC estimates for the \texttt{semi} experiment, with subpopulations ordered left to right from smallest to largest.  
Comparing estimates and uncertainties to the true population AUC, we find that the more expressive evaluation models \texttt{fixed.c} and \texttt{rand.a} form much more accurate estimates, particularly in subpoulations with smaller samples. 
\Cref{fig:nhanes-forest-comparison} further illustrates that this effect is even more pronounced when the overall subsample size is smaller.
\Cref{fig:rehosp-auc-by-subpop} shows estimates of the AUC for different subpopulations on the \texttt{dm} dataset. 
As with \texttt{semi}, the improvements are most dramatic for smaller groups; MBMs always have tighter confidence intervals than the empirical estimator, and often have more accurate point estimates.
\Cref{fig:rehosp-forest-comparison} in the appendix contains additional AUC results for other sample sizes.

For the \texttt{dm} experiment, \Cref{fig:rehosp-auc-mape} shows the mean absolute error for MBMs and the sample-based estimators as a function of the subsample size ($N$), averaging over subpopulations. 
We see a strong sample-size effect: MBMs offer the greatest improvements over the typical subsample-based estimator when sample sizes are smallest. 
\Cref{fig:app-subpop-size} contains additional results for both datasets showing performance broken down by subpopulation size.

\vspace{-.5em}
\paragraph{Model misspecification affects estimate accuracy.}
A complex but misspecified evaluation model can perform worse than a much simpler model.  
In \Cref{fig:nhanes-auc-mape} the simpler evaluation model \texttt{fixed.b} consistently outperforms the more complex \texttt{fixed.c} in the four \texttt{none} and \texttt{simple} settings. 
This makes sense, as the many pairwise interactions in \texttt{fixed.c} are not necessary to well capture the true data generating mechanism.

However, overly simple models can also perform worse than the standard empirical estimator. 
In the two \texttt{interactions} settings, the \texttt{fixed.b} model-based estimates have much higher error in estimating AUC than the sample-based estimator.
This is also clear from \Cref{fig:nhanes-auc-by-subpop}, even for some of the larger subgroups.
Fortunately, inspecting out-of-sample log-likelihoods in \Cref{fig:nhanes-nlls} identifies these as situations where the estimate should not be trusted. 
Likelihoods for the \texttt{fixed.b} model when $Y=1$ are worse than the KDE baseline, indicating that one should either iterate on the model or revert to the empirical estimator.

\vspace{-.5em}
\paragraph{Importance-weighted bootstrap confidence intervals are a faithful approximation with reasonable coverage for AUC estimates.}
\Cref{fig:ci-comparison} in the appendix compares our proposed approximate bootstrap procedure to exact bootstrapping, and shows good agreement in their respective confidence intervals.
\Cref{fig:app-coverage-tables} summarizes subpopulation estimate coverage properties.
MBMs exhibit slight overconfidence, even when using the traditional bootstrap resampling estimator, but the importance-weighted and bootstrap estimates achieve similar coverage.

\section{Discussion}
There is no substitute for better data --- validating a prediction model with limited information is a fraught exercise. 
However, in some situations we may have a modest sample size overall but limited coverage for some specific subpopulations. 
We have shown that we may be able to leverage information from other related subgroups to improve performance estimates for these smaller subpopulations.
To accomplish this task, we proposed model-based metric (MBM) estimators, a procedure to construct more accurate predictive model performance estimates for small subpopulations. 
This can offer a promising middle ground between either needing to collect additional data or being unable to draw reliable inferences about the smallest subpopulations due to extreme variance.
We found that in many settings with small subgroups, MBM's tend to have much narrower confidence intervals than the naive subsample-based estimators, and are often more accurate as well.

The difficult task of estimating ML performance for small subpopulations raises numerous issues, in both statistical methodology, fairness implications, and their intersection.
Better theoretical understanding of these model-based estimators, the use of more flexible non-linear model components, the incorporation of covariate uncertainty, and extensions to other metrics and non-classification settings all warrant further study. 

The partial pooling estimators we deploy use information across groups by design, making subgroup estimates correlated with one another.  This can potentially complicate the estimation of disparity between groups --- an area that requires further study.  Further, good theoretical coverage properties in the multi-group setting requires the development of more sophisticated estimators \citep{yu2018adaptive}.

We also adopted a simple approach for building flexible models when using continuous-valued covariates (e.g., age) --- we discretize the feature and fit a coefficient for each discrete level. There are a multitude of more sophisticated approaches for incorporating non-linear conditional models, including splines, Gaussian processes, or even neural networks. 
We used a simple class of single- and multi-level models to ensure statistical inference was reliable and efficient, but we anticipate that the incorporation of non-linear modeling components will yield additional benefits. 

Uncertainty in the covariates themselves is another potential issue.  
The evaluation model conditions on a set of demographic and additional covariates, and ignoring uncertainty in these measurements can be a source of model misspecification. 
Techniques for incorporating these uncertain measurements into the class conditional model may yield more reliable estimates of model performance.

Lastly, in this work we focused on three commonly used metrics for evaluating binary classifiers: AUC, FPR, and PPV. These metrics are purely discriminative, or rank-based; future work should also examine metrics relating to model calibration, or how well a model's predicted probabilities align with true event probabilities. Furthermore, we only applied MBMs to binary classification settings, but in principle they may also be applied for modeling other types of outcomes, e.g., categorical, continuous-valued, or survival data.

Model-based metric estimators can also be incorporated into the process of building a model card \citep{mitchell2019model} for the predictive ML model.  These estimates could help characterize the potential effects a model might have on different subpopulations, and highlight groups for which more data collection is needed.  

\paragraph{Acknowledgements.}
The authors would like to thank Lauren Hannah and Nicholas Foti for discussion of the work and input on an early draft. 

\bibliographystyle{plainnat}
\bibliography{refs.bib}

\begin{thebibliography}{27}
\providecommand{\natexlab}[1]{#1}
\providecommand{\url}[1]{\texttt{#1}}
\expandafter\ifx\csname urlstyle\endcsname\relax
  \providecommand{\doi}[1]{doi: #1}\else
  \providecommand{\doi}{doi: \begingroup \urlstyle{rm}\Url}\fi

\bibitem[Benavoli et~al.(2017)Benavoli, Corani, Dem{\v{s}}ar, and
  Zaffalon]{benavoli2017time}
Alessio Benavoli, Giorgio Corani, Janez Dem{\v{s}}ar, and Marco Zaffalon.
\newblock Time for a change: a tutorial for comparing multiple classifiers
  through {B}ayesian analysis.
\newblock \emph{The Journal of Machine Learning Research}, 18\penalty0
  (1):\penalty0 2653--2688, 2017.

\bibitem[Bosomworth(2011)]{bosomworth2011practical}
N~John Bosomworth.
\newblock Practical use of the {F}ramingham risk score in primary prevention:
  {C}anadian perspective.
\newblock \emph{Canadian Family Physician}, 57\penalty0 (4):\penalty0 417,
  2011.

\bibitem[B{\"u}hlmann(2014)]{buhlmann2014discussion}
Peter B{\"u}hlmann.
\newblock Discussion of big {B}ayes stories and {B}ayes{B}ag.
\newblock \emph{Statistical Science}, 29\penalty0 (1):\penalty0 91--94, 2014.

\bibitem[Bürkner(2017)]{brms}
Paul-Christian Bürkner.
\newblock {brms}: An {R} package for {Bayesian} multilevel models using {Stan}.
\newblock \emph{Journal of Statistical Software}, 80\penalty0 (1):\penalty0
  1--28, 2017.
\newblock \doi{10.18637/jss.v080.i01}.

\bibitem[Carpenter et~al.(2017)Carpenter, Gelman, Hoffman, Lee, Goodrich,
  Betancourt, Brubaker, Guo, Li, and Riddell]{carpenter2017stan}
Bob Carpenter, Andrew Gelman, Matthew~D Hoffman, Daniel Lee, Ben Goodrich,
  Michael Betancourt, Marcus Brubaker, Jiqiang Guo, Peter Li, and Allen
  Riddell.
\newblock Stan: A probabilistic programming language.
\newblock \emph{Journal of Statistical Software}, 76\penalty0 (1), 2017.

\bibitem[Chen and Guestrin(2016)]{Chen:2016:XST:2939672.2939785}
Tianqi Chen and Carlos Guestrin.
\newblock {XGBoost}: A scalable tree boosting system.
\newblock In \emph{Proceedings of the 22nd ACM SIGKDD International Conference
  on Knowledge Discovery and Data Mining}, KDD '16, pages 785--794, New York,
  NY, USA, 2016. ACM.

\bibitem[Dua and Graff(2017)]{Dua:2019}
Dheeru Dua and Casey Graff.
\newblock {UCI} machine learning repository, 2017.
\newblock URL \url{http://archive.ics.uci.edu/ml}.

\bibitem[D’agostino et~al.(2008)D’agostino, Vasan, Pencina, Wolf, Cobain,
  Massaro, and Kannel]{d2008general}
Ralph~B D’agostino, Ramachandran~S Vasan, Michael~J Pencina, Philip~A Wolf,
  Mark Cobain, Joseph~M Massaro, and William~B Kannel.
\newblock General cardiovascular risk profile for use in primary care.
\newblock \emph{Circulation}, 117\penalty0 (6):\penalty0 743--753, 2008.

\bibitem[Efron(1992)]{efron1992bootstrap}
Bradley Efron.
\newblock Bootstrap methods: another look at the jackknife.
\newblock In \emph{Breakthroughs in Statistics}, pages 569--593. Springer,
  1992.

\bibitem[Gelman et~al.(2013)Gelman, Carlin, Stern, Dunson, Vehtari, and
  Rubin]{gelman2013bayesian}
Andrew Gelman, John~B Carlin, Hal~S Stern, David~B Dunson, Aki Vehtari, and
  Donald~B Rubin.
\newblock \emph{Bayesian Data Analysis}.
\newblock CRC press, 2013.

\bibitem[Goutte and Gaussier(2005)]{goutte2005probabilistic}
Cyril Goutte and Eric Gaussier.
\newblock A probabilistic interpretation of precision, recall and f-score, with
  implication for evaluation.
\newblock In \emph{European Conference on Information Retrieval}, pages
  345--359. Springer, 2005.

\bibitem[Hanczar et~al.(2010)Hanczar, Hua, Sima, Weinstein, Bittner, and
  Dougherty]{hanczar2010small}
Blaise Hanczar, Jianping Hua, Chao Sima, John Weinstein, Michael Bittner, and
  Edward~R Dougherty.
\newblock Small-sample precision of {ROC}-related estimates.
\newblock \emph{Bioinformatics}, 26\penalty0 (6):\penalty0 822--830, 2010.

\bibitem[Hoffman and Gelman(2014)]{hoffman2014no}
Matthew~D Hoffman and Andrew Gelman.
\newblock The {No-U-Turn} sampler: {A}daptively setting path lengths in
  {H}amiltonian {M}onte {C}arlo.
\newblock \emph{Journal of Machine Learning Research}, 15\penalty0
  (1):\penalty0 1593--1623, 2014.

\bibitem[Huggins and Miller(2019)]{huggins2019robust}
Jonathan~H Huggins and Jeffrey~W Miller.
\newblock Robust inference and model criticism using bagged posteriors.
\newblock \emph{arXiv preprint arXiv:1912.07104}, 2019.

\bibitem[Huggins and Miller(2020)]{huggins2020robust}
Jonathan~H Huggins and Jeffrey~W Miller.
\newblock Robust and reproducible model selection using bagged posteriors.
\newblock \emph{arXiv preprint arXiv:2007.14845}, 2020.

\bibitem[Ji et~al.(2020)Ji, Smyth, and Steyvers]{ji2020can}
Disi Ji, Padhraic Smyth, and Mark Steyvers.
\newblock Can {I} trust my fairness metric? assessing fairness with unlabeled
  data and bayesian inference.
\newblock \emph{arXiv preprint arXiv:2010.09851}, 2020.

\bibitem[Johnson et~al.(2019)Johnson, Jones, and Gardner]{johnson2019gold}
Wesley~O Johnson, Geoff Jones, and Ian~A Gardner.
\newblock Gold standards are out and bayes is in: Implementing the cure for
  imperfect reference tests in diagnostic accuracy studies.
\newblock \emph{Preventive Veterinary Medicine}, 167:\penalty0 113--127, 2019.

\bibitem[Kanaya et~al.(2013)Kanaya, Kandula, Herrington, Budoff, Hulley,
  Vittinghoff, and Liu]{kanaya2013mediators}
Alka~M Kanaya, Namratha Kandula, David Herrington, Matthew~J Budoff, Stephen
  Hulley, Eric Vittinghoff, and Kiang Liu.
\newblock Mediators of atherosclerosis in south asians living in america
  (masala) study: objectives, methods, and cohort description.
\newblock \emph{Clinical cardiology}, 36\penalty0 (12):\penalty0 713--720,
  2013.

\bibitem[Lipsey and Wilson(2001)]{lipsey2001practical}
Mark~W Lipsey and David~B Wilson.
\newblock \emph{Practical meta-analysis.}
\newblock SAGE publications, Inc, 2001.

\bibitem[Mitchell et~al.(2019)Mitchell, Wu, Zaldivar, Barnes, Vasserman,
  Hutchinson, Spitzer, Raji, and Gebru]{mitchell2019model}
Margaret Mitchell, Simone Wu, Andrew Zaldivar, Parker Barnes, Lucy Vasserman,
  Ben Hutchinson, Elena Spitzer, Inioluwa~Deborah Raji, and Timnit Gebru.
\newblock Model cards for model reporting.
\newblock In \emph{Proceedings of the Conference on Fairness, Accountability,
  and Transparency}, pages 220--229, 2019.

\bibitem[Muehlematter et~al.(2021)Muehlematter, Daniore, and
  Vokinger]{muehlematter2021approval}
Urs~J Muehlematter, Paola Daniore, and Kerstin~N Vokinger.
\newblock Approval of artificial intelligence and machine learning-based
  medical devices in the {USA} and {Europe} (2015--20): {A} comparative
  analysis.
\newblock \emph{The Lancet Digital Health}, 2021.

\bibitem[{National Center for Health Statistics}(2017)]{nhanes}
{National Center for Health Statistics}.
\newblock National health and nutrition examination survey data.
\newblock {Hyattsville, MD: U.S. Department of Health and Human Services,
  Centers for Disease Control and Prevention}, 2017.

\bibitem[Oakden-Rayner and Palmer(2020)]{oakden2020docs}
Luke Oakden-Rayner and Lyle Palmer.
\newblock Docs are {ROCs}: A simple off-the-shelf approach for estimating
  average human performance in diagnostic studies.
\newblock \emph{arXiv preprint arXiv:2009.11060}, 2020.

\bibitem[Strack et~al.(2014)Strack, DeShazo, Gennings, Olmo, Ventura, Cios, and
  Clore]{strack2014impact}
Beata Strack, Jonathan~P DeShazo, Chris Gennings, Juan~L Olmo, Sebastian
  Ventura, Krzysztof~J Cios, and John~N Clore.
\newblock Impact of hba1c measurement on hospital readmission rates: analysis
  of 70,000 clinical database patient records.
\newblock \emph{BioMed Research International}, 2014, 2014.

\bibitem[Vehtari et~al.(2015)Vehtari, Simpson, Gelman, Yao, and
  Gabry]{vehtari2015pareto}
Aki Vehtari, Daniel Simpson, Andrew Gelman, Yuling Yao, and Jonah Gabry.
\newblock Pareto smoothed importance sampling.
\newblock \emph{arXiv preprint arXiv:1507.02646}, 2015.

\bibitem[Vehtari et~al.(2017)Vehtari, Gelman, and Gabry]{vehtari2017practical}
Aki Vehtari, Andrew Gelman, and Jonah Gabry.
\newblock Practical bayesian model evaluation using leave-one-out
  cross-validation and waic.
\newblock \emph{Statistics and Computing}, 27\penalty0 (5):\penalty0
  1413--1432, 2017.

\bibitem[Yu and Hoff(2018)]{yu2018adaptive}
Chaoyu Yu and Peter~D Hoff.
\newblock Adaptive multigroup confidence intervals with constant coverage.
\newblock \emph{Biometrika}, 105\penalty0 (2):\penalty0 319--335, 2018.

\end{thebibliography}

\newpage
\appendix

\section{Experimental setup details}
\label{sec:app-experimental-setup}

\subsection{Semi-synthetic NHANES experiment (\texttt{semi})}
\label{sec:app-nhanes}
We generate data using statistics from the 2017-2018 National Health and Nutrition Examination Survey, available at \url{https://wwwn.cdc.gov}.

We devise experiments in three increasingly complex data generating procedures, designed to mimic the situation where a machine learning model aims to predict cardiovascular disease risk in two categories --- low and elevated.
In all settings, we generate a population that mimics the NHANES demographic frequencies, as well as the statistics of common cardiovascular risk factors --- defined by diabetes status, hypertensive treatment status, systolic blood pressure, total cholesterol, and HDL cholesterol.

Before generating a synthetic population, we first compute statistics within each demographic category --- defined by race, age buckets, sex, and body mass index (BMI) buckets --- of cardiovascular risk factors.  Within each demographic bucket, we describe the continuous risk factors --- log systolic blood pressure, log total cholesterol, and log HDL cholesterol --- with a multivariate Gaussian distribution, and diabetes and hypertensive treatment using independent probabilities. 
To generate a unit, we first randomly draw the demographic categories from the weighted NHANES units, then draw continuous and discrete risk factors.
For each unit, we then compute the Framingham CVD risk score using the simulated risk factor values \citep{d2008general}.  We construct an outcome variable by computing the Framingham risk score, and use 10\% risk as a cutoff to define two classes --- low and elevated risk of cardiovascular disease \citep{bosomworth2011practical}.
This process results in a dataset of demographic subpopulation, covariate values, and a binary outcome, $\{ a_n, x_n, y_n \}_{n=1}^{N_pop}$, where $N_{pop} = 5e6$ units. 

Next, we construct a machine learning model score, aimed at predicting $y_n$ for each unit.  We construct such a score in three, increasingly complex ways. 

\begin{itemize} \itemsep 0pt
    \item \emph{No structure} (\texttt{none}): $S_n \sim \mathcal{N}(\mu_{y_n}, \sigma^2)$ --- the class conditional in this relationship is independent of all variables except for the true class label, $y$.  In this relationship, the AUC is identical across all subpopulations, though the false positive rate and positive predictive value can vary from subpopulation to subpopulation (due to differences in prevalence). 
     
\begin{lstlisting}[language=R]
beta.y = c(-3.5, -2.0)
S.mu = beta.y[factor(popdf$frame.class)]
S.sd = 1.25
\end{lstlisting}
    
    \item \emph{Additive structure} (\texttt{simple}): $S_n \sim \beta_a + \beta_r + \beta_g + \beta_b + \beta_y + \sigma \epsilon$ --- the class conditional mean has additive structure, based on demographic information.  This results in variation in AUC, FPR, and PPV between demographic subpopulations. 
    We set values of $\beta$ with the following R code snippet: 
    
\begin{lstlisting}[language=R]
Nr = length(unique(popdf$race))
Na = length(unique(popdf$age_bin))
Nb = length(unique(popdf$bmi_bin))
beta.gender = c(-.2, .2)
beta.race   = seq(from=-.2, to=.2, length=Nr))
beta.age    = seq(from=-.2, to=.2, length=Na))
beta.bmi    = seq(from=-.2, to=.2, length=Nb))
beta.y      = c(-3, -2.5)
S.mu = beta.gender[factor(popdf$gender)] +
       beta.race  [factor(popdf$race)] +
       beta.age   [factor(popdf$age_bin)] +
       beta.bmi   [factor(popdf$bmi_bin)] +
       beta.y     [factor(popdf$frame.class)]
S.sd = .5
\end{lstlisting}
    
    \item \emph{Interactive structure} (\texttt{interactions}): $S_n \sim \beta_{a,r} + \beta_{g,b} + \bar{f} + f_n + \sigma \cdot \epsilon$, where the age and race buckets have strong interactions, as well as the gender and BMI buckets.  Additionally, we add in the true (simulated) framingham score $f_n$, shrunken toward its global mean to mimic the compressed output of a typical machine learning classification score.  Thsi results in variation in AUC, FPR, and PPV between demographic subpopulations as well as a more difficult modeling problem. 
\begin{lstlisting}[language=R]
# interact race and age
beta.race = seq(from=-sqrt(.2), to=sqrt(.2), length=Nr)
beta.age  = seq(from=-sqrt(.2), to=sqrt(.2), length=Na)
beta.age.race = beta.age %o% beta.race
age.race.idx  = cbind(factor(popdf$age_bin), factor(popdf$race))
# interact gender and bmi
beta.gender = seq(from=-sqrt(.2), to=sqrt(.2), length=Ng)
beta.bmi    = seq(from=-sqrt(.2), to=sqrt(.2), length=Nb)
beta.gender.bmi = beta.gender %o% beta.bmi
gender.bmi.idx = cbind(factor(popdf$gender), factor(popdf$bmi_bin))
# mean value
S.mu = beta.age.race[age.race.idx] +
       beta.gender.bmi[gender.bmi.idx] +
       .7*popdf$frame.score + .3*mean(popdf$frame.score)
S.sd = .5
\end{lstlisting}
\end{itemize}

Additionally, we vary the type of class conditional noise between two settings --- homoscedastic and heteroscedastic (denoted with the suffix \texttt{-hetero}). 
In the homoscedastic setting, the conditional variance $\sigma^2$ is fixed and shared between all simulated units. 
In the heteroscedastic setting, the conditional variance is made to be a function of the conditional mean.  That function is simple sigmoid that decreases as the mean gets higher.  This makes the evaluation modeling problem much more complex, requiring us to specify distributional models that allow the residual noise variance to vary with the observed covariates. 
The heteroskedsatic error is generated with the following R code snippet:
\begin{lstlisting}[language=R]
  # heteroscedasticity
  s.fac = 1 / (1 + exp(S.mu))
  s.fac = s.fac / max(s.fac)
  S.sd  = s.fac*S.sd + .5*S.sd
\end{lstlisting}

In addition to the \emph{structure} and \emph{noise heteroscedasticity}, we also examine \emph{subsample sizes}.  That is, how accurate are estimates when only $N=1{,}000$ units are observed, vs $N=5{,}000$ units, vs $N=10{,}000$.

\subsection{Evaluation models}
\label{sec:app-evaluation-models}
For both experiments, we use a set of increasingly complex evaluation models to generate metric approximations.  The complexity incorporates additional covariates, interactions between demographic buckets, and heterscedastic noise terms. 
We fit each model using the R package \texttt{brms} \citep{brms}, drawing 4{,}000 posterior samples for each.
Additionally, for the fixed effects models we simulate 1{,}000 draws from 100 bootstrapped posteriors to generate confidence intervals. The random effects models are computationally much more expensive, so we rely on approximations of leave-one-out log-likelihood \citep{vehtari2015pareto} and the truncated importance-weighted bootstrap replicates we describe in Section~\ref{sec:approximate-bootstrap}. 

Here we present each model in \texttt{brms} formula notation (using the \texttt{bf} function for heteroscedastic models). 

\vspace{-.5em}
\paragraph{Semi-synthetic experiment evaluation models (\texttt{semi})} 
The subpopulation covariates we use (derived from NHANES) are \texttt{gender}, \texttt{race}, and \texttt{age\_bin}.  For additional covariates, we incorporate BMI, and ($\log$) diastolic and systolic blood pressure, total cholesterol, and HDL cholesterol.  We examine three fixed effects models, and two random effects models:
\begin{itemize} \itemsep 0pt
    \item fixed a: only demographic and outcome information
\begin{lstlisting}[language=R]
fixed.a = "S ~ gender + race + age_bin + bmi_bin + Y"
\end{lstlisting}
    
    \item fixed b: demographic, outcome, and additional covariate information
\begin{lstlisting}[language=R]
fixed.b = "S ~ gender + race + age_bin + bmi_bin + Y + ln.diabp + ln.ppbp + ln.tc + ln.hdl"
\end{lstlisting}

    \item fixed c: marginal and all pairwise interactions between demographic variables, and additional covariate information
\begin{lstlisting}[language=R]
fixed.c = "S ~ (gender+race+age_bin+bmi_bin+Y)^2 + ln.diabp + ln.ppbp + ln.tc + ln.hdl"
\end{lstlisting}

    \item fixed d: same as fixed c, but with heteroscedastic noise based on demographics
\begin{lstlisting}[language=R]
fixed.d = bf("S ~ (gender+race+age_bin+bmi_bin+Y)^2 + ln.diabp + ln.ppbp + ln.tc + ln.hdl",
             "sigma ~ gender + race + age_bin + bmi_bin + Y")
\end{lstlisting}

    \item random effects a: random effects based on all marginal and pairwise interctions between demographic variables, plus population level covariates
\begin{lstlisting}[language=R]
rand.a = "S ~ (1 | (gender + race + age_bin + bmi_bin + Y)^2) + ln.diabp + ln.ppbp + ln.tc + ln.hdl"
\end{lstlisting}
 
     \item random effects b: same as random effects a, but with heteroscedastic noise term based on demographic information
\begin{lstlisting}[language=R]
rand.b = bf("S ~ (1 | (gender + race + age_bin + bmi_bin + Y)^2) + ln.diabp + ln.ppbp + ln.tc + ln.hdl",
           "sigma ~ (1 | (gender + race + age_bin + bmi_bin + Y)^2)")
\end{lstlisting}
\end{itemize}

\paragraph{Rehospitalization experiment evaluation models (\texttt{dm}}
The subpopulation covariates we used are \texttt{gender}, \texttt{race}, and \texttt{age\_bin}.  
For additional covariates, we incorporate \texttt{number\_inpatient}, \texttt{number\_diagnoses},  \texttt{number\_emergency} + \texttt{CHF} and \texttt{number\_outpatient}. 
Similar to above, we examine three fixed effects models, and two random effects models:
\begin{itemize} \itemsep 0pt
    \item fixed a: only demographic and outcome information
\begin{lstlisting}[language=R]
fixed.a = "S ~ 1 + gender + race + age_bin + Y"
\end{lstlisting}
    
    \item fixed b: demographic, outcome, and additional covariate information
\begin{lstlisting}[language=R]
static.cov = " + number_inpatient + number_diagnoses + number_emergency + CHF + number_outpatient"
fixed.b = "S ~ 1 + gender + race + age_bin + Y" + static.cov
\end{lstlisting}

    \item fixed c: marginal and all pairwise interactions between demographic variables, and additional covariate information
\begin{lstlisting}[language=R]
fixed.c = "S ~ (gender + race + age_bin + Y)^2" + static.cov
\end{lstlisting}

    \item fixed d: same as fixed c, but with heteroscedastic noise based on demographics
\begin{lstlisting}[language=R]
fixed.d = bf("S ~ (gender + race + age_bin + Y)^2" + static.cov, 
             "sigma ~ gender + race + age_bin + Y"),
\end{lstlisting}

    \item random effects a: random effects based on all marginal and pairwise interctions between demographic variables, plus population level covariates
\begin{lstlisting}[language=R]
rand.a = "S ~ (1 | (gender + race + age_bin + Y)^2)" + static.cov
\end{lstlisting}
 
     \item random effects b: same as random effects a, but with heteroscedastic noise term based on demographic information
\begin{lstlisting}[language=R]
rand.b = bf("S ~ (1 | (gender + race + age_bin + Y)^2)" + static.cov,
            "sigma ~ (1 | (gender + race + age_bin + Y)^2)")
\end{lstlisting}
\end{itemize}

\section{Additional Empirical Results}

\begin{figure*}[t]
\floatconts
  {fig:app-subpop-size}
  {
  \caption{\small{Mean absolute error of AUC, FPR, and PPV estimates by subpopulation size, split into three bins, $(0, 100]$, $(100, 500]$, and $>500$.  For both the semi-synthetic and real re-hospitalization data, we observe that smaller subpopulations tend to see bigger improvements, while larger subpopulations see similar performance, averaged over the data sampling process.}}
  }
  {%
    \subfigure[\texttt{semi} experiment (interactions-hetero)]{
      \label{fig:nhanes-mape-by-size}
      \includegraphics[width=\textwidth]{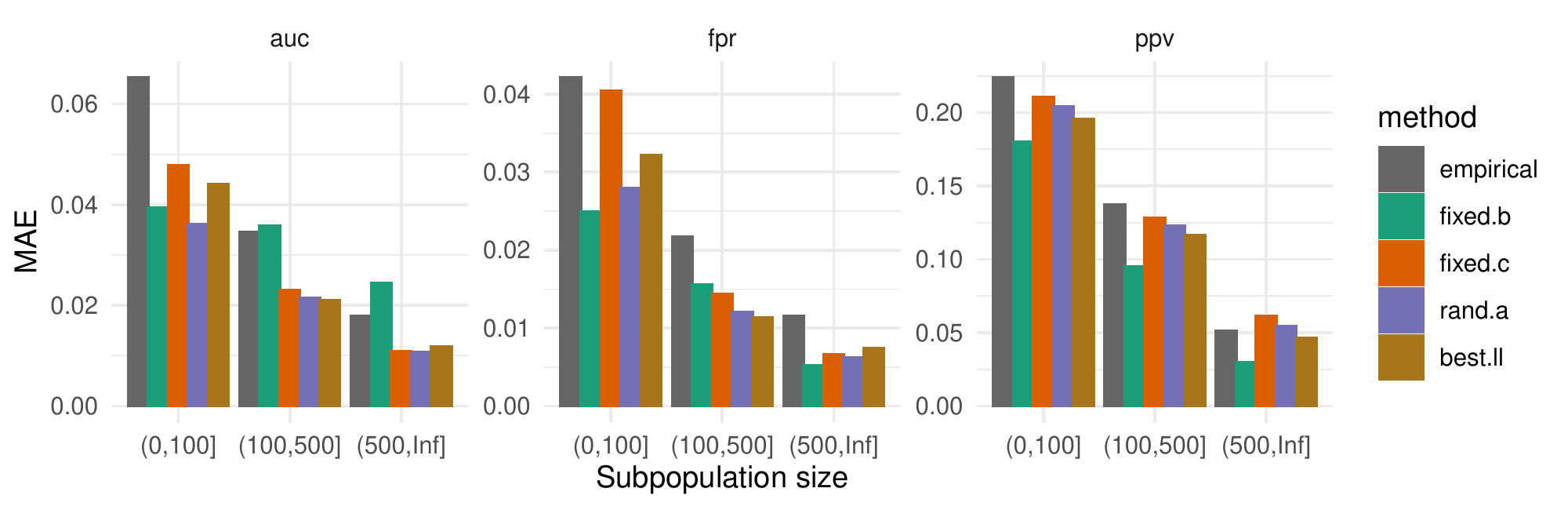}
    }
    \subfigure[\texttt{dm} experiment]{
        \label{fig:rehosp-mape-by-size}
        \includegraphics[width=\textwidth]{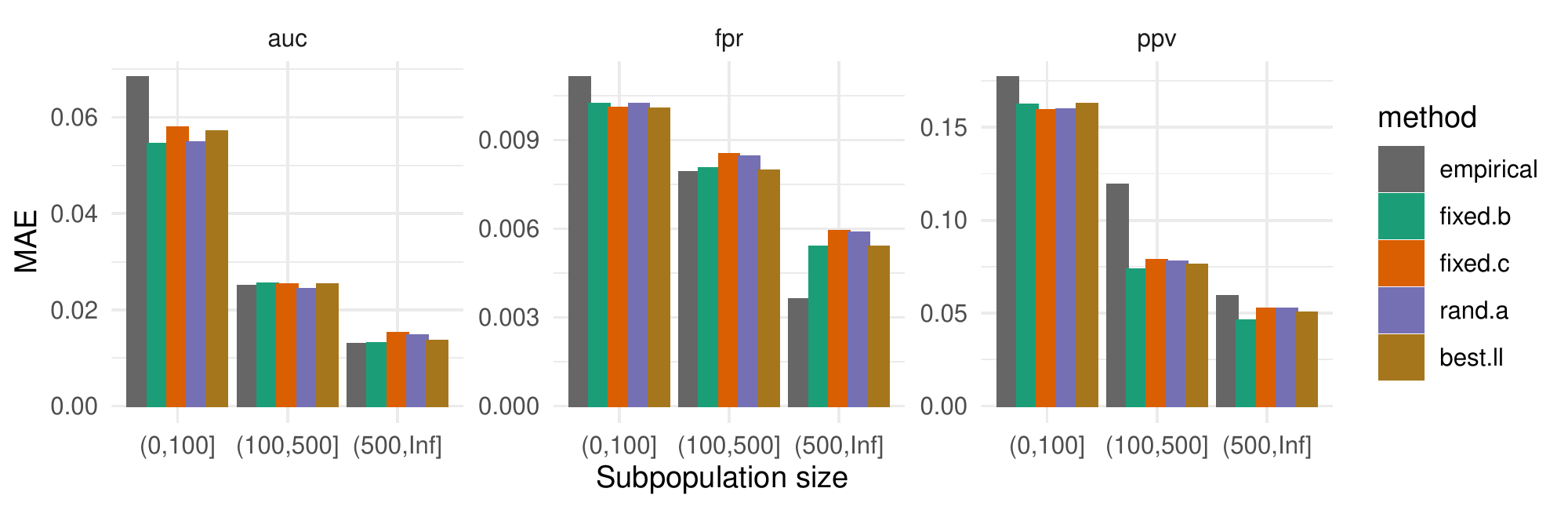}
    }
  }
  \label{fig:subpop-size}
\end{figure*}

\begin{figure*}[t]
\floatconts
  {fig:app-diabetes-forest}
  {
  \caption{\small{Comparison of AUC estimates by subpopulation for the \texttt{dm} experiment for sample sizes (a) $N=1{,}00$, (b)$N=5{,}00$, and (c) $N=10{,}00$.
  Subpopulations are sorted by size (smallest to largest). }}
  }
  {%
    \subfigure[\texttt{dm}, $N=1{,}000$]{
        \includegraphics[width=\textwidth]{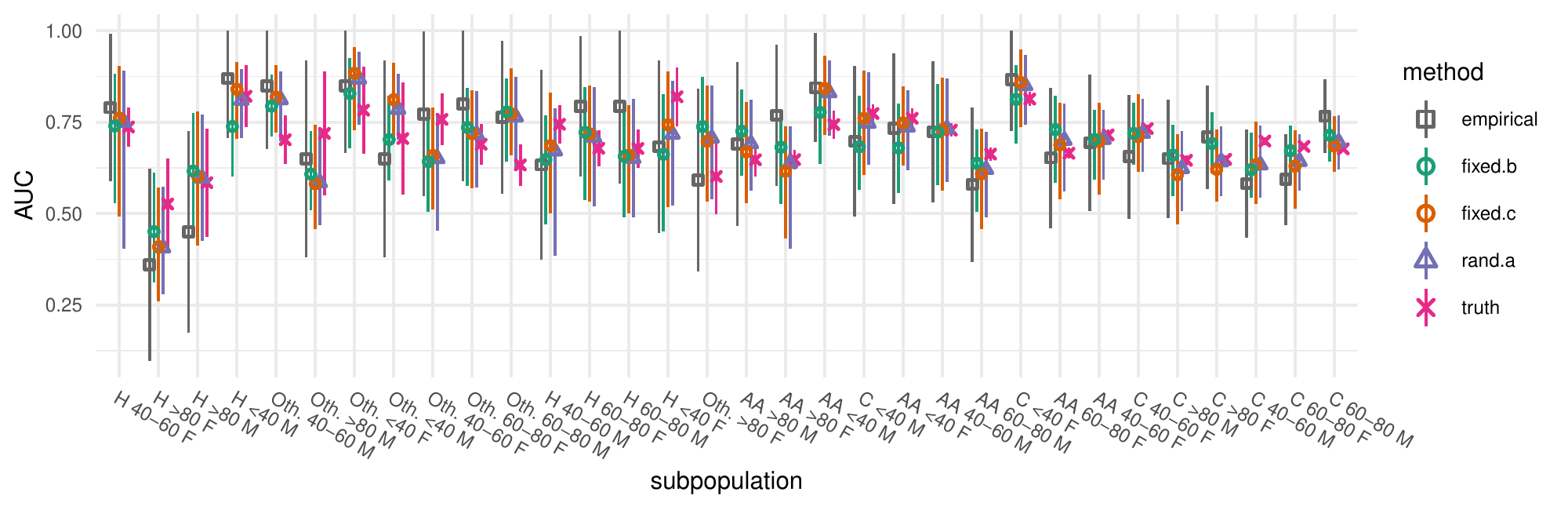}
    }
    \subfigure[\texttt{dm}, $N=2{,}000$]{
        \includegraphics[width=\textwidth]{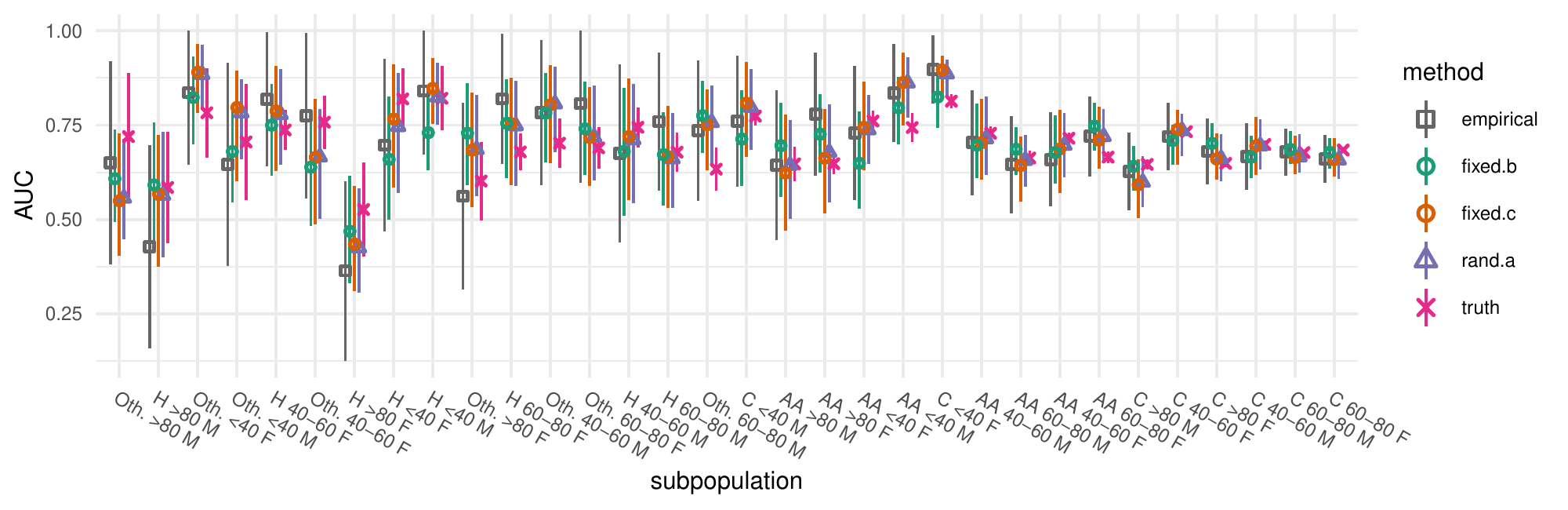}
    }
    \subfigure[\texttt{dm}, $N=5{,}000$]{
        \includegraphics[width=\textwidth]{figs/experiment-diabetes/subpop-auc-comp-diabetes-ndat=5000-run=3.pdf}
    }
  }
  \label{fig:rehosp-forest-comparison}
\end{figure*}

\begin{figure*}[t]
\floatconts
  {fig:app-nhanes-forest}
  {
  \caption{\small{Comparison of AUC estimates by subpopulation for the \texttt{semi} experiment for sample sizes for (a) $N=1{,}00$, (b)$N=5{,}00$, and (c) $N=10{,}00$.
  Subpopulations are sorted by size (smallest to largest).}}
  }
  {%
    \subfigure[\texttt{semi} \emph{interactions-hetero}, $N=1{,}000$]{
        \includegraphics[width=\textwidth]{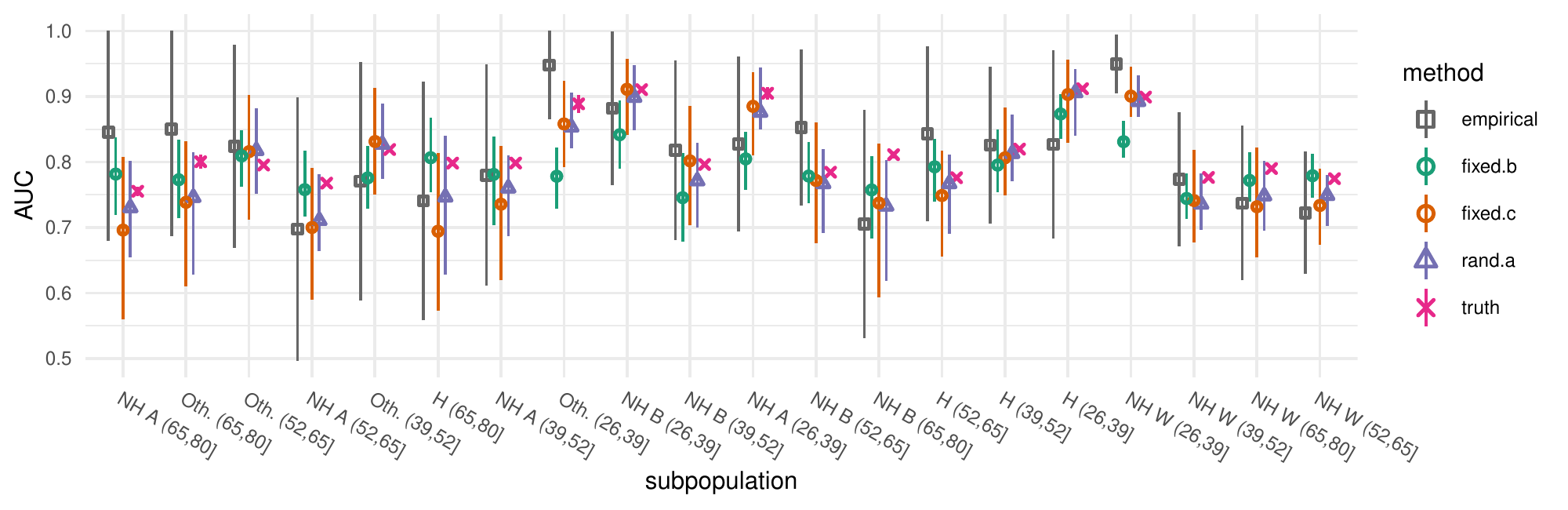}
    }
    \subfigure[\texttt{semi} \emph{interactions-hetero}, $N=5{,}000$]{
        \includegraphics[width=\textwidth]{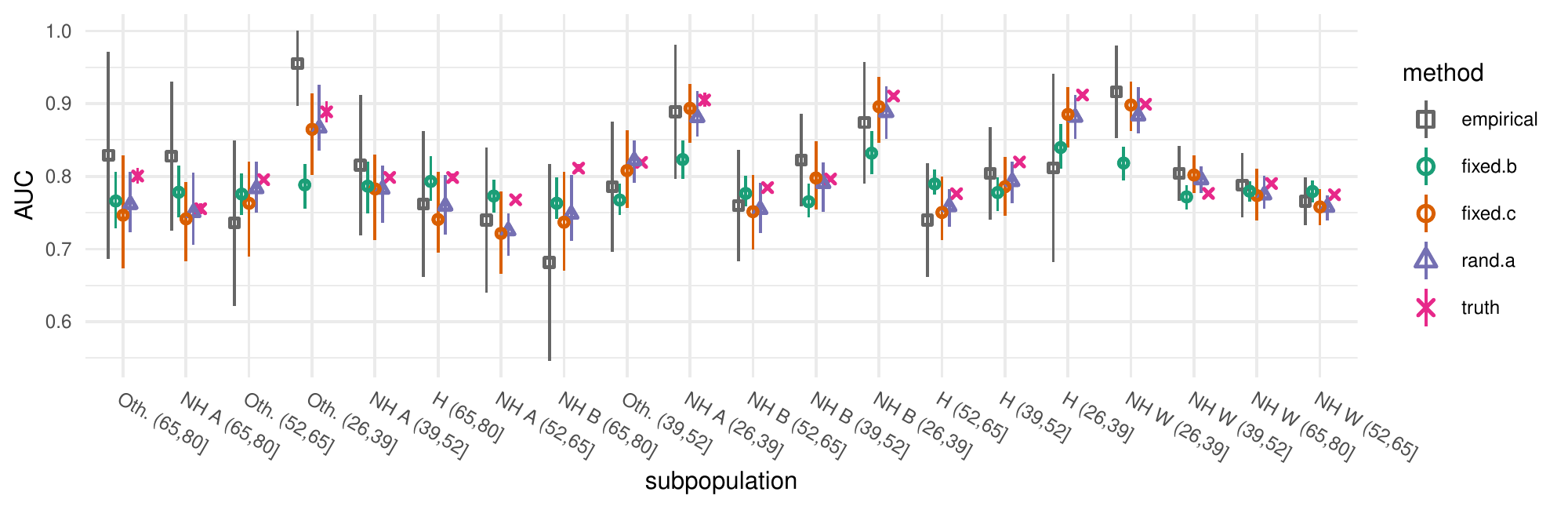}
    }
    \subfigure[\texttt{semi} \emph{interactions-hetero}, $N=10{,}000$]{
        \includegraphics[width=\textwidth]{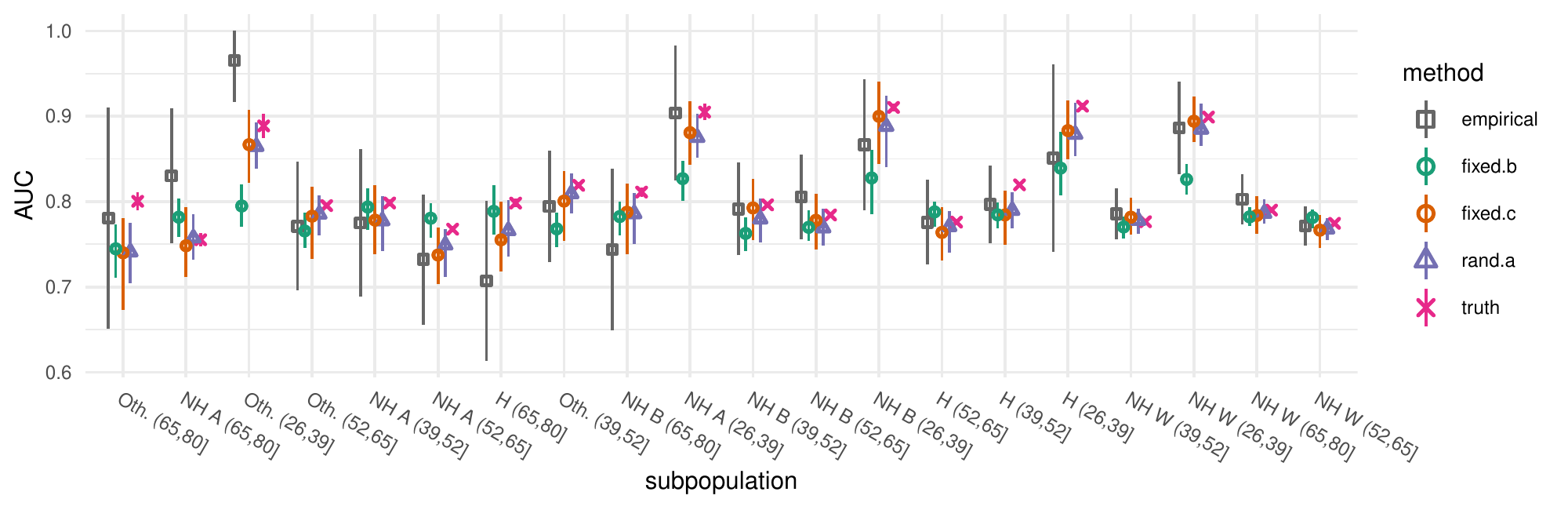}
    }
  }
  \label{fig:nhanes-forest-comparison}
\end{figure*}

\begin{figure*}[t]
\floatconts
  {fig:bootstrap-comparison}
  {
  \caption{\small{Comparison of bootstrap-estimated (slow) quantiles and approximate bootstrap-estimated (faster) quantiles, using re-weighted posterior samples.  Despite potential instability of importance weighted estimators, we see tight agreement between the two confidence interval estimators.  Note we only compare models \texttt{fixed.a}, \texttt{fixed.b}, and \texttt{fixed.c}, as the multi-level models too expensive to sample from the posterior for each bootstrap data set.}}
  }
  {%
    \subfigure[\texttt{semi}]{
      \label{fig:nhanes-ci-comparison}
      \includegraphics[width=.45\textwidth]{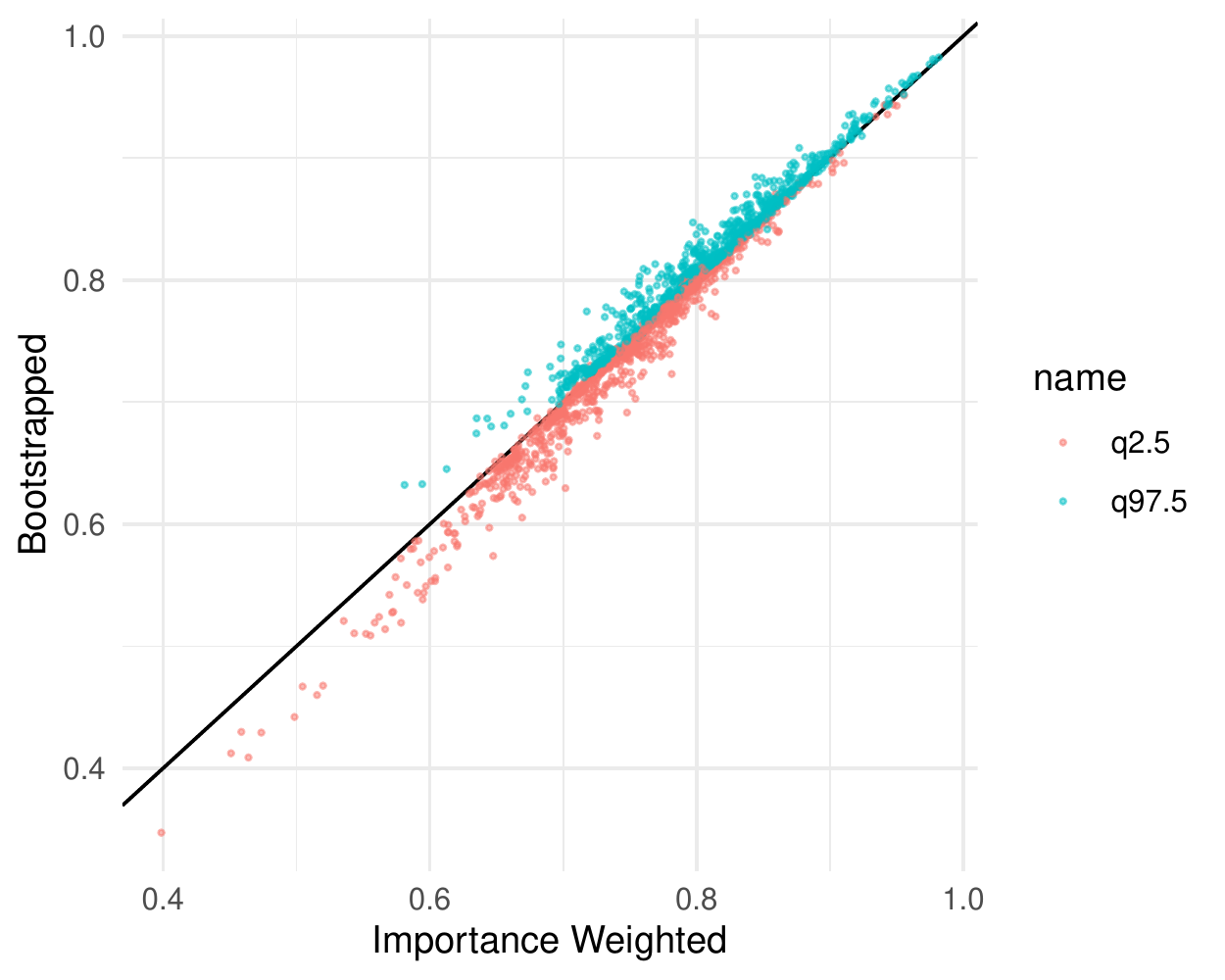}
    }
    \subfigure[\texttt{dm}]{
       \label{fig:rehosp-ci-comparison}
      \includegraphics[width=.45\textwidth]{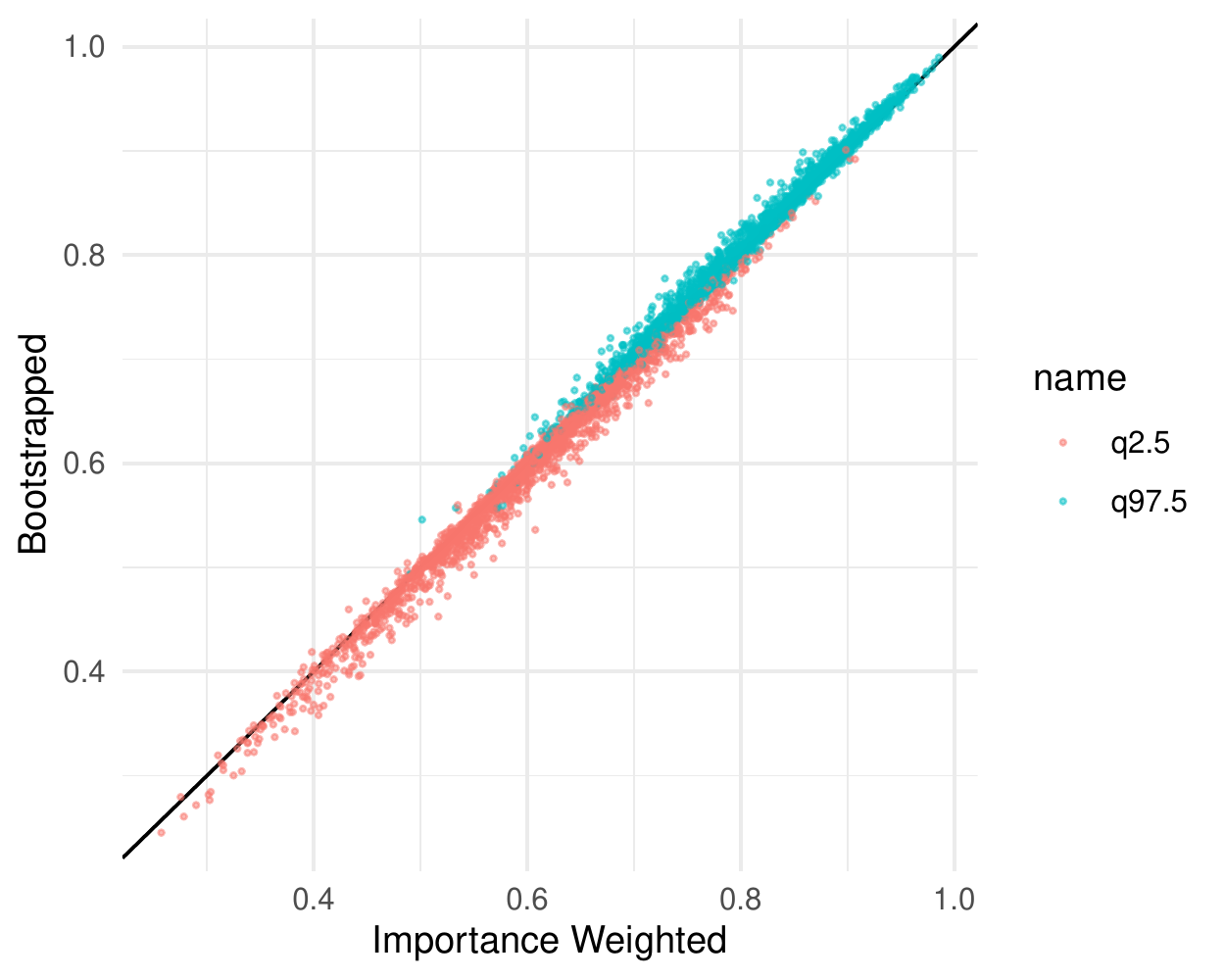}
    }
  }
  \label{fig:ci-comparison}
\end{figure*}

\begin{figure*}[t]
\floatconts
  {fig:app-coverage-tables}
  {
  \caption{\small{Empirical coverage of 50\% and 95\% confidence intervals for AUC, FPR, and PPV, as well as the total range over all (exact or approximate) bootstrap samples.   
  The empirical estimator is reported first.
  We compare the importance weighted (iw) and full bootstrap intervals for evaluation models \texttt{fixed.a}, \texttt{fixed.b}, and \texttt{fixed.c}.  For evaluation models \texttt{fixed.d}, \texttt{rand.a}, and \texttt{rand.b} we only compute the more efficient importance weighted estimate.  We find the \texttt{empirical} estimator using the bootstrap to be well-calibrated for AUC, but overly certain for FPR and PPV.  We find that MBMs are also overly certain, but only slightly more or equally so, while also producing more accurate estimates on average.}}
  }
  {%
    \subfigure[\texttt{semi} coverage (\emph{interactions-hetero})]{
        \begin{tabular}{lrrr|rrr|rrr}
         \toprule
          & \multicolumn{3}{c}{AUC} & \multicolumn{3}{c}{FPR} & \multicolumn{3}{c}{PPV} \\
          method & 50\% & 95\% & range & 50\% & 95\% & range & 50\% & 95\% & range \\ 
         \midrule
         \input{figs/experiment-semi-synth/wide-coverage-table-nhanes-interactions-hetero-all.tex}
         \bottomrule
        \end{tabular}
        \label{tab:semi-coverage}
    }
    \vspace{2em} \\
    \subfigure[\texttt{dm} coverage]{
        \begin{tabular}{lrrr|rrr|rrr}
         \toprule
          & \multicolumn{3}{c}{AUC} & \multicolumn{3}{c}{FPR} & \multicolumn{3}{c}{PPV} \\
          method & 50\% & 95\% & range & 50\% & 95\% & range & 50\% & 95\% & range \\ 
         \midrule
         \input{figs/experiment-diabetes/wide-coverage-table-diabetes.tex}
         \bottomrule
        \end{tabular}
        \label{tab:rehosp-coverage}
    }
  }
  \label{tab:coverage}
\end{figure*}

\end{document}